\documentclass{article}
\usepackage{log_2025}
\usepackage{float}
\usepackage[normalem]{ulem}
\usepackage{multirow}
\usepackage{graphicx} 
\usepackage{algorithm}
\usepackage{algpseudocode}
\usepackage{booktabs}						
\usepackage{multirow}						
\usepackage{amsfonts}						
\usepackage{graphicx}						
\usepackage{duckuments}						

\usepackage[numbers,compress,sort]{natbib}	

\def \R{\mathbb{R}}
\def \beq{\begin{equation}}
\def \eeq{\end{equation}}
\def \ba{\begin{array}}
\def \ea{\end{array}}

\def \X{\mathcal{X}}
\def \V{\mathcal{V}}
\def \E{\mathcal{E}}
\def \G{\mathcal{G}}

\title[DyGSSM]{DyGSSM: Multi-view Dynamic Graph Embeddings with SSM Gradient Update}

\author[Pijani et al.]{%
Bizhan Alipour Pijani\\
University of North Texas \\
\email{pijanialipourpijani@my.unt.edu}\And
Serdar Bozdag\\
University of North Texas\\
\email{serdar.bozdag@unt.edu}
}

\begin{document}

\maketitle

\begin{abstract}

Dynamic graphs whose topology and nodes evolve over time are ubiquitous in multiple real world domains such as social networks, finance, and healthcare. Traditional graph learning methods fail to capture structural changes and temporal patterns in dynamic graphs. Recent advances in dynamic graph representation learning, such as meta-learning-based approaches, have addressed some of these challenges. However, existing methods still face three key limitations. First, most approaches capture either local or global structures of the graphs, neglecting to model both simultaneously. Second, meta-learning models often depend on user-specific window size, which must be carefully tuned for each dataset. A short window size may miss trends, and a long window size may blur recent updates.  Third, most methods work on only discrete-time or continuous-time dynamic graphs, resulting in suboptimal performance across different temporal settings. 
 To address these limitations in dynamic graph representation learning, we propose a novel method called DyGSSM (Multi-view Dynamic Graph Embeddings with SSM Gradient Update). We extract local and global features at each snapshot and fuse them using a lightweight attention mechanism for link prediction. To capture long-term dependencies when updating model parameters, we incorporate HiPPO (High-order Polynomial Projection Operators) algorithm, which has gained attention for its ability to efficiently optimize and preserve sequence history in State Space Models (SSMs). 
 DyGSSM is designed to handle both discrete-time and continuous-time dynamic graphs. Parameter comparisons show that DyGSSM requires substantially fewer parameters than the other methods. Extensive experiments on 12 public datasets demonstrate that DyGSSM outperforms baselines and State-Of-The-Art (SOTA) methods in 32 out of 36 evaluation metrics. The source code and datasets are available at https://github.com/bozdaglab/DyGSSM.
\end{abstract}
\section{Introduction}\label{sec:Introduction}
Dynamic graphs are ubiquitous as many real-world interactions and relationships are dynamic in nature, such as social networks~\cite {pareja2020evolvegcn}, transportation networks \cite{cini2023scalable}, transaction networks \cite{shamsi2022chartalist}, and trade networks \cite{poursafaei2022towards}.
 Unlike static graphs, dynamic graphs evolve over time, with changes occurring in their topology or edge/node attributes. Message passing-based graph representation learning models \cite{kipf2016semi,velivckovic2017graph,ozdemir2024igcn,kesimoglu2023supreme,hamilton2017inductive} have achieved significant success in graph analysis tasks. These models are effective in capturing local structural information due to the inherent locality of the message-passing mechanism. These models have recently been extended to dynamic graphs for representation learning \cite{you2022roland, zhu2023wingnn,wu2020temp,zhao2019t,li2024dynamic}. 
   For example, in \cite{zhao2019t,bonner2019temporal,xue2021modeling} ``snapshot'' of the dynamic graph at each time point is processed using a graph representation learning approach and sequence encoders such as Long Short Term Memory (LSTM) \cite{pareja2020evolvegcn} or Transformers \cite{sankar2020dysat} are utilized to capture the temporal evolution of the graph over time. Although these approaches have shown promising results, they have low expressive power, as an independent GCN is trained for each snapshot of the graph. As a result, these models may fail to extract historical structural information. To address this, researchers have integrated the sequence encoder into the GCN layer to update the parameters of the GCN model \cite{pareja2020evolvegcn,sankar2020dysat} over time. However, these methods have another limitation. In dynamic graphs, changes occurring in multiple hops away from a source node could still influence the source node in subsequent snapshots. However, these methods fail to extract signal from such distant nodes, thereby reducing their expressive power. 

Meta-learning has emerged as an effective approach for modeling temporal dynamics. In meta-learning, which is based on the idea of model transfer, the model parameters are updated based on previous time steps and then passed to the next time steps. In this setting, ROLAND \cite{you2022roland} updated node embeddings generated by the GNN layer 
utilizing adjacent time snapshots. 
However, this approach only aggregates two adjacent snapshots and neglects temporal information with long-term dependencies. 
 WinGNN \cite{zhu2023wingnn} introduced a sliding window approach to update the model parameters.
WinGNN is mainly designed for discrete-time snapshots and may not handle irregularly timed events. 
 It relies on several hyperparameters, such as window size, beta value, and meta-learning rate, which require careful tuning for each dataset. Additionally, if the window size is too small, it may miss trends, and if too large, it may blur recent updates.  
 Moreover, these methods are GNN-based and typically focus on local neighbors (local view), while ignoring the global structure of the graph (global view). Both views provide complementary information: local views capture fine details of immediate interactions, while global views capture long-range changes that may affect future states. Long-distance information is especially important in time series data, as dynamic graphs evolve over time and interactions can occur at any step \cite{wang2023networked,miller2023balancing}. Changes in distant hops may also influence the source node in later snapshots. Therefore, it is important to extract both local and global features, which together can be regarded as a multi-view representation of the same snapshots.

Recently, state space models (SSMs) have become a popular and powerful tool for sequence modeling. 
 Some SSM-based methods have been proposed for dynamic graphs \cite{ding2024dygmamba,yuan2024dg}. For example, GraphSSM \cite{li2024state} leverages SSMs to capture continuous-time dynamics in dynamic graphs for node classification tasks. DyGMamba \cite{ding2024dygmamba} uses two types of SSM, a node- and time-level SSM. 
  DyGMamba encodes one-hop temporal neighbors of nodes; as a result, it may miss higher-order structural dependencies. Additionally, the node-level SSM still sequentially processes interaction histories for each node. In very dense graphs or when histories are extremely long, this sequential processing can become computationally expensive and slow. As shown in the results section, dynamic SSM-based models often require a large number of parameters, which makes them less practical for large datasets. 

To address these challenges, we introduce Multi-view Dynamic Graph Embeddings with State Space Model Gradient Update (DyGSSM). While traditional SSMs model time-series data directly, DyGSSM leverages SSM to update model parameters. To initialize the SSM state, we utilized High-order Polynomial Projection Operators (HiPPO) \cite{gu2022train,gu2021efficiently}. 
 To make this process computationally tractable, we introduce a compressed parameter-space representation: instead of flattening every scalar weight into the SSM state, we treat each parameter tensor as a single entity, maintaining one HiPPO state per tensor. This drastically reduces memory and computational overhead and reduces the number of hyperparameters (such as window size). The SSM in DyGSSM considers the loss of each snapshot during parameter updates, which facilitates smoother updates and introduces a mechanism for 
\textit{forgetting} less relevant information from the past time 
 while \textit{remembering} critical patterns from earlier ones.  
  This design enables DyGSSM to encode both discrete-time (snapshot-based) and continuous-time dynamic graphs. 
  Additionally, DyGSSM introduces a lightweight attention-fusion mechanism that departs from conventional multi-head attention. 
  We summarize our main contributions as follows:
\begin{itemize}
    \item To the best of our knowledge, we are the first to integrate SSM into the meta-learning strategy to update model parameters. We introduce an SSM-based method to effectively capture long-term dependencies when updating model parameters. This approach avoids the need for numerous hyperparameters, which can increase the model’s sensitivity to specific datasets.  
    \item  We introduce a compressed HiPPO formulation that maintains one SSM state per parameter tensor rather than per-weight, which significantly reduces computational and memory costs.
    
    \item DyGSSM is designed to handle both discrete-time (snapshot-based) and continuous-time dynamic graphs, extending its applicability to a broader range of temporal graph scenarios. 

    \item We introduce a lightweight, and scalable attention mechanism that fuses local and global embeddings efficiently.
    
    \item Extensive experiments on multiple datasets demonstrate the superiority of DyGSSM over State-Of-The-Art (SOTA) models, while having the lowest number of parameters.
\end{itemize}
\section{Related Work}\label{sec:RelatedWorks}
 Dynamic graph representation models can broadly be categorized into the following three groups. 
\subsection{Sequence-Based Models}\label{sec:SequenceBased}
Sequence-based models follow the message passing and temporal encoder to capture time dependencies \cite{wu2020temp}.  Researchers have utilized GCN with RNN variants to capture time dependencies. 
 For example, CD-GCN \cite{manessi2020dynamic} is a combination of GCN and LSTM. They apply GCN to obtain the embeddings of each snapshot and pass the embeddings to LSTM for time sequence dependencies.  
EvolveGCN \cite{pareja2020evolvegcn} and GC-GCN-N \cite{jiang2022graph} integrate GCN and GRU for tasks such as link prediction, edge classification, node classification, and landslide displacement forecasting. GC-GCN-N captures spatial dependencies among monitoring stations through a weighted adjacency matrix and temporal patterns from time-series data using GRU.  PoGeVon  \cite{wang2023networked} introduces an encoder-decoder architecture for dynamic graph representation. They utilize a novel node position embedding derived from the random walk with restart (RWR) approach. In addition, they use the concept of the sliding window with a Lagrange multiplier to control the amount of information that can be transmitted through the latent representation. They use a 2-layer GRU to capture the dynamic information in networked time series. 
 These models have two main limitations. First, they do not share parameters across time steps; instead, each time step trains an independent GCN, which restricts the model’s ability to leverage historical structural information. Second, they require a large number of parameters because they rely on sequence-based models (i.e., GRU and LSTM) to capture the graph’s temporal evolution.
\subsection{Meta-Learner-Based Models}\label{sec:MetaLearnerBased}
Meta-learning is based on the idea of transfer learning, where previous experience is used to quickly adapt to a new task. In dynamic graphs, meta-learning–based models extend static GNNs by learning model parameter initializations for the next time steps. 
 ROLAND \cite{you2022roland} 
 extends static graphs to dynamic ones with minimal extra computational cost. They use a two-layer GNN, where each layer updates its parameters and passes them to the adjacent snapshot. WinGNN \cite{zhu2023wingnn} 
  proposes a framework that combines GNN with a meta-learning strategy and a novel random gradient aggregation mechanism. Instead of relying on temporal encoders, WinGNN models graph dynamics by introducing a randomized sliding-window strategy that computes loss on each snapshot and propagates updated model parameters to the next snapshot. They perform backpropagation only at the end of the window. 
    MetaDyGNN \cite{yang2022few}  leverages a meta-learning strategy for few-shot link prediction in dynamic graphs. They introduce time interval-wise and node-wise adaptations to encompass time dependencies and node dependency features, and update the global parameters. 
 These models suffer from one main limitation. The meta-learning parameters, such as the meta-learning rate and window size, must be carefully tuned for each dataset, which adds extra complexity. For example, short window size (or a ROLAND-based parameter update) may fail to capture long-term trends, while a long window size may obscure recent updates.
\subsection{SSM- and Transformers-based methods}
 Many researchers have used transformers instead of LSTM or GRU to capture the temporal evolution of dynamic graphs \cite{jin2023spatio}. For example, Dysat \cite{sankar2020dysat} employs self-attention in two different aspects. First, attending to structural neighborhoods at each time point. Second, attending to previous historical representations to conduct link prediction. 
  Graph Transformers (GT) have gained popularity in the field of graph representation \cite{biparva2024todyformer,yu2023towards,ye2022meta}. 
   For example, TransformerG2G \cite{varghese2024transformerg2g} utilizes transformer for learning temporal graphs. They use only transformer encoder to learn intermediate node representations from all the previous snapshots up to the current snapshot. They use two projection heads (linear mapping and non-linear mapping) to generate low dimensional latent embedding at different snapshots. 
   DTFormer \cite{chen2024dtformer} 
   collects all the first-hop neighbors of source and destination nodes. Then, it maps these neighbor features into a sequence to be processed by transformer.  Despite the effectiveness of transformer on graph-structured data, it suffers from having a quadratic computational cost and lack of inductive biases on graph structures. 
   
    Recent successes of SSM-based models (such as Mamba) in computer vision and natural language processing tasks have motivated researchers to apply SSM-based models to graphs. For example, DyGMamba \cite{ding2024dygmamba} introduces two levels of SSM: a node-level SSM to encode node interactions and a time-level SSM to exploit temporal patterns. DG-Mamba \cite{yuan2024dg} treats a dynamic graph as a self-contained system, using an SSM to capture global intrinsic dynamics. It discretizes the system state according to cross-snapshot graph adjacency, enabling the model to capture long-range dependencies through a selective snapshot scanning strategy.
 Dyg-mamba \cite{li2024dyg} proposes a new continuous SSM for dynamic graph learning. They consider irregular time spans
as control signals for SSM to have robust and generalizable model. 
 Although these methods achieve good performance, their high computational and memory costs make it difficult to scale them to large dynamic graphs. Moreover, none of the SSM-based models integrate SSM and meta-learning to address the limitations of meta-learning while enabling the model to distinguish which information to forget and which information to remember from the past.
\section{Preliminaries}\label{Preliminaries}
In this section, we introduce the notion of a discrete and continuous-time dynamic graph and other important components that are adopted in DyGSSM.
\subsection{Problem Formulation}
Let $\V$ be a set of nodes and $\E$ be the set of edges that connect the nodes in $\V$. 
A graph consists of three components $\G = (\V, \E, \X)$, where $\X\in\R^{n\times m}$ is a node feature matrix, $n=|\V|$ and $m$ is the dimension size of the feature. For a graph $\G$, we can create an adjacency matrix $A \in \R^{n\times n}$, that represents local neighbors of each node
as follow: 
\beq 
\mathcal{A}_{ij} = \left\{
\ba{ll}
1 & \mbox{ if } (v_i, v_j) \in \E, \\
0 & \mbox{ otherwise. }
\ea
\right.
\eeq
To study a discrete-time dynamic graph, we let $G =\{\G_1,\dots,\G_T\}$ be a sequence of graphs for discrete snapshots $t =1, \dots, T$. Here, each $\G_t = (\V_t , \E_t , \X_t )$ represents a snapshot of the dynamic graph with adjacency matrix $\mathcal{A}_t$ at time $t$. 
 The local neighbors of a node $i$ at time $t$ denoted as 
 $\mathcal{N}_{t, i}^{local} = \{  v_j \mid (v_i, v_j) \in \mathcal{E}_t \}$
 and the global neighbors of a node $i$ at time $t$ denoted as 
 $\mathcal{N}_{t, i}^{global} = \{ v_j \mid v_j \in \mathcal{RW}_t(v_i) \}$
 where $\mathcal{RW}_t(v_i)$ denotes the set of nodes visited by random walk ($\mathcal{RW}$) starting from $v_i$ in snapshot $t$.
In contrast, a continuous-time dynamic graph models interactions as an event stream. We represent the graph as a sequence of non-decreasing chronological interactions  
\[
G = \{(u_1,v_1,t_1), (u_2,v_2,t_2), \dots\},
\]  
with $0 \leq t_1 \leq t_2 \leq \dots$, where $u_i, v_i \in \V$ denote the source and destination nodes of the $i$-th interaction at timestamp $t$. Each node $u \in \V$ may be associated with a feature vector $x_u \in \mathbb{R}^{d_N}$, and each interaction $(u,v,t)$ may carry an edge feature $e^t_{u,v} \in \mathbb{R}^{d_E}$, where $d_N$ and $d_E$ denote the dimensions of the node feature and link feature. 
To evaluate DyGSSM, we consider a link prediction task in discrete- and continuous-time dynamic graphs.  The model takes the local ($\X_t^{local}$) and global ($\X_t^{global}$) node embeddings of nodes $v_i$ and $v_j$ in discrete/continuous time, and outputs the probability of a connection between them in the next time frame or event.
\subsection{State Space Model}
A SSM frames a discrete-time system by a linear mapping from a discrete input $u_t$ at time $t$ to a discrete output $y_t$ through a state variable $s_t$ and three matrices, namely $K$, $B$, and $C$ as follows:
\begin{align}
    s_t &= K s_{t-1} + B u_t
    \label{eq:ssm_discrete_state} \\
    y_t &= C s_t 
\end{align}

The structure of the matrix $K$ is important when building an SSM, as this matrix determines which part of the previous state can be passed to the current state.  In continuous time, the hidden state ($s(t)$) evolves as follows:
\begin{align}
    \frac{d s(t)}{dt} &= K s(t) + B u(t) \label{eq:ssm_cont_state} \\
    y(t) &= C s(t)
\end{align}
where $u(t)$ is a continuous input signal and $y(t)$ is the corresponding output.  
 To use the SSM for meta-learning-based parameter updates without relying on a window size, we start from Equation~\ref{eq:ssm_discrete_state} and modify it slightly to obtain
\begin{equation} \label{hipssm}
    s_t = \hat{K} s_{t-1} + \hat{G}_t \cdot weight_t
\end{equation}
where $s_{t-1}$ is the state vector at time $t-1$, initialized to zero, $\hat{G}_t$ is the gradient vector of the model parameters at time $t$, $weight_t$  is the reciprocal of the loss (see Section \ref{snapshotoptim}), and $\hat{K}$ is the projection matrix. 
 Using HiPPO to initialize the matrix $K$ was shown to perform better than initializing it as a random matrix. The  HiPPO matrix is designed to generate a hidden state that can memorize the past inputs with no tunable hyperparameters. We initialized $\hat{K}$ using the HiPPO algorithm \cite{gu2022train,gu2021efficiently} as follow: 
\begin{equation}
\hat{K}_{i, j} =
    \begin{cases}
      (-1) ^{i-j} (2i + 1) & \text{if i > j}\\
      2 & \text{if i= j}\\
      0 & \text{otherwise}
    \end{cases} 
\end{equation}
For entries where the row index is greater than the column index ($i> j$), the equation above fills the values along the lower diagonal. $(2i +1)$ acts as linear function of the row index $i$ and takes a positive or negative sign depending on whether $i - j$ in $(-1)^{i - j}$ is even or odd. If $i - j$ is even, the output is positive; otherwise, it is negative. On the diagonal ($i= j$), this matrix has a value of 2, and above the diagonal ($i< j$), the values are set to 0. The Equation \ref{hipssm} in continuous time becomes as follows:
\begin{equation} \label{hipssmcon}
    s(t) = \hat{K} s(t) + \hat{G}(t) \cdot weight(t)
\end{equation}

\section{Method}\label{sec:Methods}
In this section, we describe the DyGSSM architecture. We start by explaining how we compute local and global node embeddings. Next, we show how these two types of embeddings are combined. Finally, we discuss how the model optimizes its parameters using the HiPPO method. Fig.~\ref{fig:modelarc} shows DyGSSM architecture and Fig.~\ref{fig:enter-arc} (see Appendix~\ref{appbaseline}) represents the parameter update mechanisms in ROLAND, WinGNN, and DyGSSM. Note that we only show local and global representations in the discrete-time setting here, but the same process can be applied to the continuous-time case as well. 
\begin{figure}
    \centering
    \includegraphics[width=0.55\linewidth]{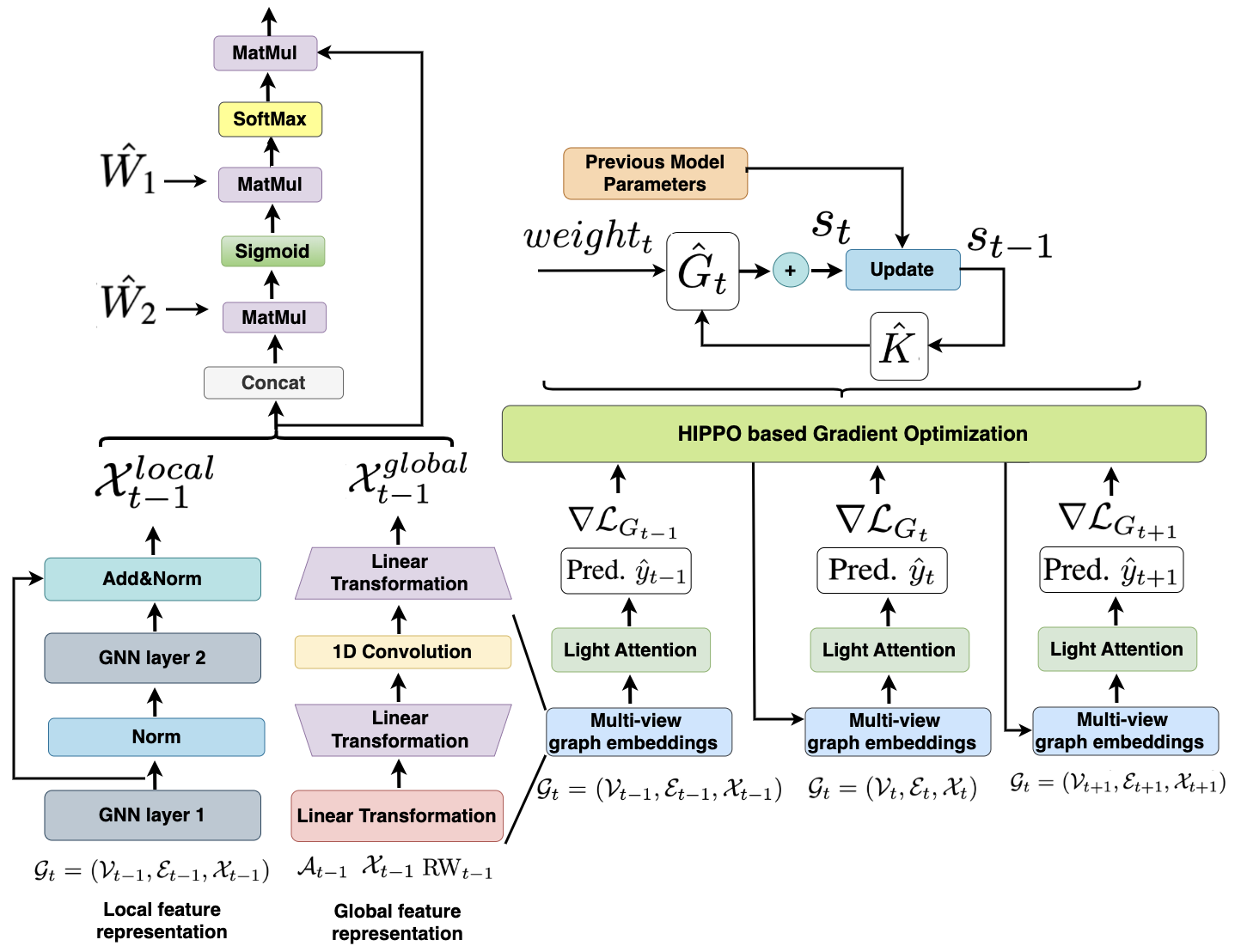}
    \caption{DyGSSM architecture. }
    \label{fig:modelarc}
\end{figure}
\subsection{Node Embeddings}\label{sec:GAT}
Here, we discuss local and global node embeddings. 
\subsubsection{Local Embeddings}
Graph representation techniques update the embedding of each node $u$ in $\G_t$ by performing message passing between neighbor nodes as follows:
\beq\label{eq:gcn}
h^{(l)}_{t,u} = UPDATE(h^{(l - 1)}_{t,u} , AGG({h^{(l - 1)}_{t,v}, \forall v \in \mathcal{N}^{local}_{t, u}})),
\eeq
where \( h^{(l)}_{t,u} \), the embedding of node \( u \) at layer \( l \), is computed by aggregating information from all its local neighbors at time point \( t \), denoted by \( \mathcal{N}^{local}_{t,u} \). This process can be expressed in matrix form as $H^{(l)}_t = \sigma \left( \hat{A}_t H^{(l-1)}_t W^{(l)}_t \right)$
where \( H^{(l)}_t \in \mathbb{R}^{n \times d} \) represents the embeddings of all nodes in \( \mathcal{G}_t \) at layer \( l \), $\hat{A}_t = D_t^{-\frac{1}{2}} A_t D_t^{-\frac{1}{2}}$
is the normalized adjacency matrix with \( A_t \) being the adjacency matrix of \( \mathcal{G}_t \) and \( D_t \) being the corresponding diagonal degree matrix. Here, \( W^{(l)}_t \) is the learnable weight matrix of the \( l \)-th layer at time \( t \), and \( \sigma \) denotes a non-linear activation function.
 We used two-layer message passing to compute local node embeddings ($\X^{local}_t$). 
 \subsubsection{Global Embeddings}
 Given a snapshot of a graph at time $t$, $\G_t$, to compute  global embeddings for nodes, we computed a biased $\mathcal{RW}$-node embedding as follows: For each node in $\G_t$, we customized the $\mathcal{RW}$ to explore far-away nodes from the source node. 
 To make sure that we are not selecting an arbitrary global neighbor for each node, we ran the biased $\mathcal{RW}$ for each node 50 times with a path length of 5. This results in a list of 250 nodes for each source node. We selected the five most frequent nodes in the sequence for each source node. 
  
  To increase the efficiency of the training phase, we precomputed the global neighbors of each node before the training phase. However, $\mathcal{RW}$ can still be costly on large graphs as the biased $\mathcal{RW}$ must be performed for each node.  Alternatively,  $\mathcal{RW}$ can be recalculated for the entire graph only when new data arrives, which can increase the overall inference time. To mitigate these cost-prohibitive scenarios, we introduced a caching mechanism that stores node edges from prior snapshots. At each snapshot, we identify newly introduced nodes or those whose topology has changed, and recompute the neighbors for only those nodes. 
   For the remaining snapshots, we reuse the cached distributions to avoid redundant computation.  This would allow the model to reuse previously computed information and update embeddings incrementally, rather than recomputing them entirely, which reduces $\mathcal{RW}$ computational cost (see Fig. \ref{fig:snapshot_times_scaled} in Appendix~\ref{modelscalability}).  After computing the $\mathcal{RW}$ for the source node, we obtain a sequence of five nodes, including the source node \( i \) as follows, \( \mathcal{RW}^i_t = [v^i_t, v^1_t, v^2_t, v^3_t, v^4_t] \). The embeddings of the generated sequence are passed through a linear transformation:
\[
z^i_t = W_1 \, \mathcal{RW}^i_t + b_1,
\]
where \( W_1 \) and \( b_1 \) are learnable parameters that expand the embedding dimension. Next, a 1D convolution is applied to generate embeddings for each source node:
\[
e^i_t = Conv1D(z^i_t),
\]
where \( Conv1D(\cdot) \) denotes the 1D convolution operation. Finally, another linear transformation reduces the dimension back to the original size:
\[
x^{global}_t = W_2 e^i_t + b_2,
\]
with learnable parameters \( W_2 \) and \( b_2 \). The final global node embeddings for all nodes in \( \mathcal{G}_t \) are denoted as \( \X^{global}_t \). Note that this global node embedding approach is highly parameter-efficient compared to GRU, LSTM, or transformer encoders.

\subsection{Integration of
Local and Global Node Embeddings}\label{sec:CA}
To fuse the local and global embeddings, we first  concatenate $\X^{local}_t$ and $\X^{global}_t$ as follows:
\beq \label{f8}
h^{concat}_t = concat(\X^{local}_t,  \X^{global}_t)
\eeq
Next, we compute attention weights for each embedding as follows:
\begin{equation}
a_t = \text{softmax} \left( \hat{W_{\text{1}}} \cdot \sigma(\hat{W_{\text{2}}} \cdot h^{concat}_t) \right)
\label{eq:attention}
\end{equation}
where $\hat{W}_1$ and $\hat{W}_2$ are two learnable attention matrices with shapes $2 \times d \times d_a$ and $2 \times d_a \times 2$, respectively. Here, $d_a$ denotes the size of attention weights computed for each embedding, and the 2 corresponds to the number of embeddings (local and global). $a_t$ acts as a gate that controls the amount of information that $\X^{local}_t$ and $\X^{global}_t$ can transmit to the final representation of the graph at time $t$. This single lightweight gating operation has no head count or depth parameters. As a result, the performance is not sensitive to some parameters (e.g., the number of attention layers).  The final representation is given as follows:
\begin{equation}
h^{fused}_t = a_t \cdot  h^{concat}_t 
\label{eq:final_rep}
\end{equation}
\subsection{Gradient Optimization using HiPPO}\label{snapshotoptim}
Consistent with existing research, we use the cross-entropy loss as our loss function: 
\beq    \label{eq:loss}
\mathcal{L}_t = - \frac{1}{M} \sum_{(u, v) \in \E_t}^M y^t_{u, v} \cdot log (\hat{y}^t_{u, v}) + (1 - y^t_{u, v}) \cdot log(1 - \hat{y}^t_{u, v})
\eeq
where $ \hat{y}^t_{u, v}$ is calculated as follow
\[
\hat{y}^t_{u, v} = MLP(concat(h^{fused}_{u,t}, h^{fused}_{v,t}))
\]
The main objective of this step is to calculate the gradient of the current snapshot $t$ with respect to $\mathcal{L}_t$, efficiently optimize model parameters, and propagate it to the next snapshot. 
To do that, we first calculate the reciprocal of the loss, acting as a simple dynamic weighting mechanism to adjust the influence of each snapshot during parameter updates, as presented below:
\begin{equation}\label{eq9}
    weight_t = \frac{1}{\mathcal{L}_t + \epsilon} 
\end{equation}
This dynamic weighting helps the model prioritize parameter updates from snapshots where it performs well (i.e., when the loss $\mathcal{L}_t$ is small), while reducing the influence of updates from snapshots with poor performance.  The small constant $\epsilon$ prevents division by zero and ensures numerical stability.
 We followed Equation \ref{hipssm} to update the SSM state, and performed model parameter updates as follows 

\begin{align}\label{modelupdate}
    loss\_scale = \min(loss, max\_loss\_scale) \\
    gate = tanh(\Theta_t \times s_t) \\
    scaled\_gate = clamp(gate, -\ max\_gate, max\_gate) \\
    \Theta_{t+1} \leftarrow \Theta_{t} -loss\_scale \times scaled\_gate
\end{align}

In the update equation, \( \Theta_{t} \) denotes the model parameter in the current snapshot, and \( \Theta_{t + 1} \) is the updated model parameter for the snapshot $t+1$. The loss value capped by a constant value, $max\_loss\_scale$, to produce the loss scaling factor $loss\_scale$, which prevents excessively large gradients or NaN values during training.  
  The raw gradients ($\Theta_{t + 1}$) are then elementwise multiplied by the SSM state ($s_t$), passed through a hyperbolic tangent function to bound their range, and then clamped to the interval (\([- max\_gate, max\_gate]\)), producing $scaled\_gate$ value.  Finally, the update step multiplies this loss scaling factor ($loss\_scale$) by the scaled gate ($scaled\_gate$) and subtracts the result from the current parameters ($\Theta_{t}$), ensuring stable and bounded parameter updates. 
  In our implementation, we 
   set $max\_gate=1.0$ and $max\_loss\_scale=0.1$. Both constants are non-learnable and were not subject to hyperparameter tuning. These two factors aim to mitigate instability caused by spiky or large loss values. 

   In continuous dynamic graphs, each interaction (edge event) is processed sequentially according to its timestamp rather than aggregated into fixed snapshots. When a new event ($u$, $v$, $t$) arrives, DyGSSM uses the cached $\mathcal{RW}$ representation and runs $\mathcal{RW}$ only for the affected nodes, ensuring that computation scales linearly with the number of active nodes. The HiPPO-based SSM state maintains a continuous temporal memory between irregular events, updating the parameter trajectory which preserves long-range dependencies even under uneven event intervals. This design allows DyGSSM to operate efficiently in event-driven streams without reconstructing full snapshots or retraining the model. 
\section{Experiments}\label{sec:results}
\textit{\textbf{Datasets.}} We evaluated DyGSSM on 12 different datasets, commonly employed in dynamic graph representation studies. All datasets are available on the SNAP website (https://snap.stanford.edu/) and DyGFormer paper.  
 We provide detailed descriptions of each dataset in Appendix~\ref{appdata}. To ensure a fair comparison with existing methods, we follow two settings in the discrete dynamic graph. 
  In the WinGNN setting, we used the evaluation code (https://github.com/pursuecong/WinGNN.git)
 provided by the WinGNN authors, while in the HawkesGNN \cite{qi2025input} setting, we used the evaluation code (https://github.com/oncemoe/hawkesGNN.git)
 provided by the HawkesGNN authors. For example, WinGNN employed 1000 negative samples when computing MRR, whereas HawkesGNN used 100. For continuous dynamic graphs, we followed the evaluation code (https://github.com/yule-BUAA/DyGLib.git)
 released with DyGFormer. In all cases, we adopted the preprocessing and data splits from the respective repositories. 
\textit{\textbf{Baselines.}} To verify the superiority of DyGSSM, we compared its results with various recent dynamic graph models on the link prediction task in discrete-time dynamic graphs, including EvolveGCN,
DGNN \cite{manessi2020dynamic}, dyngraph2vec \cite{goyal2020dyngraph2vec}, ROLAND, WinGNN, TransformerG2G \cite{varghese2024transformerg2g}, DTFormer \cite{chen2024dtformer}, DySAT \cite{sankar2020dysat}, VGRNN \cite{hajiramezanali2019variational}, HTGN \cite{yang2021discrete}, M2DNE \cite{lu2019temporal}, GHP \cite{shang2019geometric}, HawkesGNN and DG-Mamba \cite{yuan2024dg}, and continuous-time dynamic graphs, including JODIE \cite{kumar2019predicting}, DyRep \cite{trivedi2019dyrep}, TGAT \cite{xu2020inductive}, TGN \cite{rossi2020temporal}, CAWN \cite{wang2021inductive}, EdgeBank \cite{poursafaei2022towards}, TCL \cite{wang2021tcl}, GraphMixer \cite{cong2023we}, DyGFormer \cite{yu2023towards}, and FreeDyG \cite{tian2024freedyg}. We did not compare DyGSSM with GraphSSM because GraphSSM focuses on node classification, which differs from our task. 
 We explain each of the baseline methods in Appendix~\ref{appbaseline}. We describe the evaluation metrics and implementation details in Appendix~\ref{app-eval-metrics}. 
\subsection{Link Prediction Results}
Table \ref{tab:reddit_title_metrics} presents the link prediction results for DyGSSM and SOTA models in WinGNN settings (see Table~\ref{table:dataset} in  Appendix~\ref{appdata} for the dataset statistics). In this table, the results for EvolveGCN-H, EvolveGCN-O, DGNN, Dyngraph2vec, ROLAND, and WinGNN were taken directly from the WinGNN paper, while we ran the remaining models using the same train/test split and random seed. 
As shown, DyGSSM outperformed SOTA models in 18 out of 20 cases with substantial improvements. For instance, on Bitcoin-Alpha, DyGSSM boosted accuracy to 92.33\% and AUC to 96.71\%, far exceeding the second best baselines. Similar trends were observed on DBLP and Reddit-Title, where DyGSSM achieved high accuracy (97.63\% and 99.79\%, respectively) and substantially higher MRR and Recall@10 compared to the second best methods. On the UCI dataset, DyGSSM delivered the best performance across all metrics, improving accuracy to 96.82\%, AUC to 98.56\%, and Recall@10 to 58.43\%, consistently surpassing competing models. On Bitcoin-OTC, while DTFormer and WinGNN achieved the highest AUC and Recall@10, respectively, DyGSSM slightly improved accuracy  while substantially improving MRR. Importantly, across datasets where older baselines struggled with scalability or memory issues, DyGSSM did not encounter Out-Of-Memory (OOM) errors and consistently provided large performance margins, particularly in MRR and Recall@10. 
 A key observation is that in datasets with large snapshots,  such as Bitcoin-Alpha and Reddit-Title, the improvement was promising. This suggests that as the dataset gets bigger in terms of snapshots, SSM can effectively capture long-range dependencies in those datasets. 
\begin{table*}[ht]
    \caption{Link prediction performance comparison on five datasets. The best and second best results are shown in \textbf{bold} and \underline{underlined}, respectively. 
     We repeated the experiment with 10 random seeds and reported the average metrics with standard deviation. The * indicates that due to memory constraints, the  number of negative samples was reduced from 1000 to 50. OOM indicates that OOM occurred when we attempted to run the model in our environment, even with smaller negative samples.}
    \resizebox{\textwidth}{!}{%
    \begin{tabular}{@{}lccccccccccccc@{}}
        \toprule
        \textbf{Dataset} & \textbf{Metric} & \textbf{ EvloveGCN-H} & \textbf{ EvloveGCN-O} &\textbf{DGNN}& \textbf{dyngraph2vec}& \textbf{ROLAND} & \textbf{WinGNN} & \textbf{TransformerG2G}& \textbf{DTFormer}&\textbf{DG-Mamba}& \textbf{DyGSSM} \\ \midrule
        \multirow{4}{*}{Bitcoin-Alpha} 
&Accuracy&51.99±0.2546&57.44±0.4096&OOM&OOM&66.21±2.7566& \underline{81.17±0.5058}&OOM&80.44±0.0238 &OOM& \textbf{92.33$\pm$0.0013}\\
        & AUC&63.71±1.0318 &68.93±0.9144&OOM&OOM&90.21±1.1762&91.43±0.3259&OOM&\underline{95.62±0.0174}&OOM&  \textbf{96.71$\pm$0.0276}\\
        & MRR&3.28±0.2845&2.52±0.1014&OOM&OOM&14.52±0.6506&\underline{36.74±3.9389}&OOM&*&OOM&\textbf{62.97$\pm$0.0281}\\
        & Recall@10&7.06±1.1900&5.27±0.5093&OOM&OOM&31.25±2.2782&\underline{64.55±3.6126}&OOM&*&OOM& \textbf{88.69$\pm$0.0618}
        \\ \bottomrule
        \multirow{4}{*}{Bitcoin-OTC} 
        & Accuracy&50.48±0.0321&50.56±1.5719&54.08±0.6755&58.29±4.5547&86.60±0.5233&\underline{87.14±1.2408}&0.5$\pm$0.0000&77.49$\pm$0.0266&OOM& \textbf{88.45$\pm$0.0027}\\
        & AUC&55.38±1.6617 &59.82±2.5744&59.13±6.4914&62.12±10.7457&90.07±1.2998&91.64±0.6178&58.43$\pm$0.0594&\textbf{97.59$\pm$0.0034}&OOM&\underline{94.37$\pm$0.0478}\\
        & MRR&11.27±0.5793 &11.44±0.4986&15.16±0.5773&35.39±2.5046&16.54±1.2191&\underline{37.94±1.7019}&*&*&OOM& \textbf{52.49$\pm$0.0165}\\
        & Recall@10&20.58±1.6515&26.40±2.1204 &31.09±2.1594&58.29±6.7410&41.77±3.3926&\textbf{73.96±1.4569}&*&*&OOM& \underline{70.47$\pm$0.0281}
        \\ \bottomrule
        \multirow{4}{*}{DBLP} 
        & Accuracy&63.17±0.4138&65.24±0.5294 &OOM&OOM&62.87±0.5908&68.43±0.4135&49.99±0.0000&\underline{70.07±0.0130}&51.97±0.0211 & \textbf{97.19$\pm$0.0002} \\
        & AUC&70.91±0.3823 &72.64±0.4697&OOM&OOM&77.79±0.1689&\underline{77.87±0.3050}&53.01±0.00946&77.80±0.0167&52.17±0.0293 & \textbf{99.37$\pm$0.0005}\\
        & MRR&2.55±0.0032& 2.48±0.0038&OOM&OOM&6.60±0.0047&\underline{7.46±0.0020}&3.42±0.0028&*&* & \textbf{27.90$\pm$0.0449}\\
        & Recall@10&5.12±0.0310 &4.84±0.0023&OOM&OOM&13.48±0.0132&\underline{16.63±0.0299}&6.87±0.0046 &*&*& \textbf{62.11$\pm$0.0732}
        \\ \bottomrule
        \multirow{4}{*}{Reddit-Title}               
&Accuracy&85.85±0.0164&77.46±1.2696&OOM&OOM&93.42±0.0073&\underline{99.55±0.0009}&OOM&82.74$\pm$0.0008&OOM&\textbf{99.79$\pm$0.0004}\\
& AUC&93.87±0.0054& 97.17±0.2683&OOM&OOM&97.90±0.0001&\underline{99.87±0.0002}&OOM&94.25$\pm$0.0004&OOM&\textbf{99.99$\pm$0.0000} \\
& MRR&3.28±0.0198& 1.31±0.0213&OOM &OOM&\underline{35.11±0.0928}&29.91±0.0829&OOM&*&OOM&\textbf{66.62$\pm$0.0211}\\
& Recall@10& 5.05±0.6796&1.81±0.2453&OOM&OOM&\underline{61.13±0.0970}&60.46±0.2910&OOM&*&OOM&\textbf{97.47$\pm$0.0205}
        \\ \bottomrule
       \multirow{4}{*}{UCI}           
&Accuracy&59.85±2.5388&49.91±1.4492&50.91±0.0510&50.88±3.1146&81.83±0.6433&\underline{86.70±1.1867}&50.00$\pm$0.0000&78.84±0.0160&63.52±0.0040& \textbf{96.82$\pm$0.0148}\\
& AUC&71.99±1.8252& 62.05±3.8124& 52.19±0.5604&54.30±1.1352&91.81±0.3052&\underline{94.05±0.4679}&65.32$\pm$0.0809&87.10±0.0149&58.88±0.0040& \textbf{98.56$\pm$0.0093}\\
& MRR&8.17±0.2284&10.81±0.5327 & 1.52±0.0016&17.84±0.4917&11.84±0.2561&\underline{21.69±0.3383}&17.46$\pm$0.0422&19.36±0.0769&17.14±0.0095& \textbf{25.95$\pm$0.0570}\\
& Recall@10&14.37±0.4915 &16.94±0.9584 &4.56±0.7313&36.22±1.6716&25.14±0.9237&\underline{40.62±0.9364}&29.92$\pm$0.0487&29.30±0.1027&28.91±0.0108 & \textbf{58.43$\pm$0.1529}
        \\ \bottomrule
        
    \end{tabular}}
    \label{tab:reddit_title_metrics}
\end{table*}
 
  
  Table~\ref{tab:overall_performance_HawkesGNN} shows the comparison of DyGSSM with SOTA models in HawkesGNN settings (see  Table~\ref{HawkesGNNsettings} in Appendix~\ref{appdata} for the dataset statistics). We obtained all the results from the HawkesGNN paper and ran DyGSSM under the same settings. As presented, DyGSSM outperformed all the models on five out of seven datasets. Particularly, on the Reddit-Title and Reddit-Body datasets, DyGSSM outperformed the second-best model by 17\%. 
 We also compared DyGSSM with models designed for continuous-time dynamic graphs under both inductive and transductive settings. As shown in Table~\ref{cont_dynamicinap} (see  Table~\ref{tab:dataset_stats} in Appendix~\ref{appdata}   for the dataset statistics), DyGSSM outperformed all competing models across five datasets in the transductive and inductive setting with respect to Average Precision (AP). Particularly, DyGSSM outperformed the second-best method by about 50\% in UN Trade dataset both in inductive and transductive settings. We also compared the AP of our model with several baseline models on the discrete datasets, as presented in Table \ref{tab:ap} (see Appendix~\ref{resultsappendix}). As shown, our model outperformed other models except on Bitcoin-OTC dataset.  
 
 We encountered an OOM issue (Table~\ref{tab:reddit_title_metrics}) when sampling 1000 negative edges (following WinGNN settings) to compute MRR and Recall@10 for DTFormer, DGMamba, and TransformerG2G. To address this, we reduced the number of negative samples to 50 edges for these calculations. We observed that DyGSSM outperformed all the models on DBLP, Reddit-Title, and Bitcoin-Alpha (see Table~\ref{tab:50mrr} in Appendix~\ref{resultsappendix}). On Bitcoin-OTC, DTFormer achieved better MRR and Recall@10, while DyGSSM was the second-best model.  If a model still encounters an OOM issue there, we report it as OOM in Table \ref{tab:reddit_title_metrics}; otherwise, we mark it with $*$. 
  Finally, Table \ref{tab:auc_merged} (see Appendix~\ref{resultsappendix}) presents comparisons of AUC for transductive and inductive continuous-time dynamic graph link prediction tasks, where DyGSSM again consistently outperformed baselines. 
\begin{table}[!ht]
\centering
\small
\caption{Overall performance (MRR@100) comparison on seven datasets. Each experiment was conducted using three random seeds, and the average performance is reported along with the standard error.}
\label{tab:overall_performance_HawkesGNN}
\resizebox{\textwidth}{!}{%
\begin{tabular}{lcccccccc}
\toprule
\textbf{Methods} & \textbf{Bitcoin-OTC} & \textbf{Bitcoin-Alpha} & \textbf{UCI} & \textbf{Reddit-Title} & \textbf{Reddit-Body} & \textbf{AS733} & \textbf{StackOverflow} \\
\midrule
DySAT        & 21.39 $\pm$ 2.79 & 19.16 $\pm$ 2.21 & 23.31 $\pm$ 9.42 & 17.46 $\pm$ 4.18 & 13.87 $\pm$ 3.90 & 25.10 $\pm$ 1.71 &  OOM \\
EvolveGCN   & 7.84 $\pm$ 0.09  & 6.65 $\pm$ 0.55  & 7.33 $\pm$ 0.15  & 30.67 $\pm$ 0.00 & 18.55 $\pm$ 0.02 & 42.06 $\pm$ 0.00  & 31.21 $\pm$ 0.48 \\
Roland      & 30.94 $\pm$ 0.70 & 32.97 $\pm$ 1.78 & 17.04 $\pm$ 2.30 & 46.33 $\pm$ 0.27 & 38.57 $\pm$ 0.42 & 21.21 $\pm$ 5.73 &  38.57 $\pm$ 1.44 \\
WinGNN      & 3.86 $\pm$ 1.26  & 3.90 $\pm$ 0.84  & 2.37 $\pm$ 0.13  & 4.19 $\pm$ 1.25  & 2.69 $\pm$ 0.38  & 4.29 $\pm$ 2.10  &  7.51 $\pm$ 0.67 \\
VGRNN       & 6.62 $\pm$ 0.10  & 6.49 $\pm$ 0.29  & 6.96 $\pm$ 0.08  & OOM              & 17.19 $\pm$ 0.14 & 41.94 $\pm$ 2.04 &  OOM \\
HTGN        & 6.36 $\pm$ 0.06  & 7.72 $\pm$ 0.66  & 8.67 $\pm$ 0.43  & 11.50 $\pm$ 0.98 & 10.70 $\pm$ 0.52 & 13.86 $\pm$ 0.58 &  OOM \\
GraphMixer  & 43.67 $\pm$ 0.25 & 35.72 $\pm$ 0.41 & 33.63 $\pm$ 0.02 & 38.32 $\pm$ 0.01 & 33.15 $\pm$ 0.02 & 28.86 $\pm$ 0.00 &  OOM \\
M2DNE       & 7.82 $\pm$ 1.05  & 5.49 $\pm$ 0.29  & 8.86 $\pm$ 0.44  & 5.40 $\pm$ 0.05  & 6.03 $\pm$ 0.38  & 19.43 $\pm$ 0.12           & OOM \\
GHP         & 3.40 $\pm$ 0.41  & 3.40 $\pm$ 0.46  & 4.15 $\pm$ 0.14  & 16.00 $\pm$ 2.32 & 8.33 $\pm$ 2.00  & 22.15 $\pm$ 4.88 &  OOM \\
Hawkes-GCN  & \underline{46.16 $\pm$ 0.45} & \textbf{47.87 $\pm$ 5.85} & 35.61 $\pm$ 0.06 & 47.44 $\pm$ 0.20 & 36.44 $\pm$ 0.42 & 44.34 $\pm$ 0.41 &  46.41 $\pm$ 0.31 \\
Hawkes-GAT  & \textbf{51.34 $\pm$ 0.07} & \underline{40.66 $\pm$ 0.25} & \underline{35.59 $\pm$ 1.58} & \underline{50.84 $\pm$ 0.05} & \underline{40.97 $\pm$ 0.47} & \underline{45.95 $\pm$ 0.79} &  \underline{48.83 $\pm$ 0.14} \\
\bottomrule
\\
DyGSSM  & 35.75$\pm$0.00 & 30.22 $\pm$ 0.00& \textbf{36.08 $\pm$ 0.03} & \textbf{59.38 $\pm$ 0.00} & \textbf{48.04 $\pm$ 0.00} & \textbf{52.64 $\pm$ 0.00} & \textbf{52.43$\pm$0.00} \\
\bottomrule
\end{tabular}}
\end{table}
\begin{table*}[!ht]
\centering
\caption{Average Precision (AP) for transductive and inductive dynamic link prediction with random negative sampling strategies. The best and second best results are shown in \textbf{bold} and \underline{underlined}, respectively. The results are taken from the DyGFormer and FreeDyG papers. Since the FreeDyG authors did not evaluate their model on Can. Parl, US Legist, and UN Trade, we used the results reported for DyGFormer on these datasets and marked FreeDyG with “–”. Inductive results for EdgeBank were not reported by either DyGFormer or FreeDyG and are marked as “–”. . }
\resizebox{\textwidth}{!}{%
\begin{tabular}{l|ccccccccccccc}
\toprule
\textbf{Settings} & \textbf{Datasets} & \textbf{JODIE} & \textbf{DyRep} & \textbf{TGAT} & \textbf{TGN} & \textbf{CAWN} & \textbf{EdgeBank} & \textbf{TCL} & \textbf{GraphMixer} & \textbf{DyGFormer} & \textbf{FreeDyG} & \textbf{DyG-Mamba} &\textbf{DyGSSM} \\
\midrule
\multirow{5}{*}{\rotatebox[origin=c]{90}{Transduction}} & Enron       & 79.10$\pm$0.85 & 82.02$\pm$3.07 & 72.58$\pm$0.79 & 85.33$\pm$1.05 & 89.56$\pm$0.09 & 83.53$\pm$0.00 & 79.70$\pm$0.71 & 81.08$\pm$0.73 & 92.47$\pm$0.12 & 92.51$\pm$0.05 & \underline{93.14$\pm$0.08} & \textbf{94.64 ± 0.01}  \\
 & UCI         & 87.65$\pm$1.85 & 70.24$\pm$0.32 & 79.55$\pm$0.83 & 90.69$\pm$0.45 & 94.35$\pm$0.11 & 76.20$\pm$0.00 & 88.12$\pm$2.73 & 93.50$\pm$0.49 & 95.76$\pm$0.15 & \underline{96.28$\pm$0.11} & 96.14$\pm$0.14&\textbf{98.95 $\pm$ 0.00}\\
 & Can. Parl.  & 69.26$\pm$0.31 & 66.54$\pm$2.76 & 70.73$\pm$0.72 & 70.88$\pm$2.34 & 69.82$\pm$2.34 & 64.55$\pm$0.00 & 68.67$\pm$2.67 & 77.04$\pm$0.46 & 97.36$\pm$0.45 & - & \underline{98.20$\pm$0.52}& \textbf{99.99 $\pm$ 0.00}  \\
 & US Legis.   & 75.05$\pm$1.52 & 75.34$\pm$0.39 & 68.52$\pm$3.16 & 75.99$\pm$0.58 & 70.58$\pm$0.48 & 58.39$\pm$0.00 & 69.59$\pm$0.48 & 70.74$\pm$1.02 & 71.11$\pm$0.59 & - & \underline{73.66$\pm$1.13} &\textbf{92.91 $\pm$ 0.04}\\
 & UN Trade    & 64.94$\pm$0.31 & 63.21$\pm$0.93 & 61.47$\pm$0.18 & 65.03$\pm$1.37 & 65.39$\pm$0.12 & 60.41$\pm$0.00 & 62.21$\pm$0.03 & 62.61$\pm$0.27 & 66.46$\pm$1.29 & - &\underline{68.51$\pm$0.17} &\textbf{99.99 $\pm$ 0.00}\\
\bottomrule
\multirow{5}{*}{\rotatebox[origin=c]{90}{Inductive}} &
Enron       & 80.72 $\pm$ 1.39 & 74.55 $\pm$ 3.95 & 67.05 $\pm$ 1.51 & 77.94 $\pm$ 1.02 & 86.35 $\pm$ 0.51 &- &76.14 $\pm$ 0.79 & 75.88 $\pm$ 0.48 & 89.76 $\pm$ 0.34 &89.69 ± 0.17& \underline{91.14 ± 0.07}&\textbf{97.58 ± 0.00}\\
& UCI         & 79.86 $\pm$ 1.48 & 57.48 $\pm$ 1.87 & 79.54 $\pm$ 0.48 & 88.12 $\pm$ 2.05 & 92.73 $\pm$ 0.06 &- & 87.36 $\pm$ 2.03 & 91.19 $\pm$ 0.42 & 94.54 $\pm$ 0.12 &\underline{94.85 ± 0.10}& 94.15 ± 0.04& \textbf{97.06 ± 0.00}\\
& Can. Parl.  & 53.92 $\pm$ 0.94 & 54.02 $\pm$ 0.76 & 55.18 $\pm$ 0.79 & 54.10 $\pm$ 0.93 & 55.80 $\pm$ 0.69 &- & 54.30 $\pm$ 0.66 & 55.91 $\pm$ 0.82 & 87.74 $\pm$ 0.71 &-&\underline{90.05 $\pm$ 0.86}& \textbf{99.99$\pm$ 0.00}\\
& US Legis.   & 54.93 $\pm$ 2.29 & 57.28 $\pm$ 0.71 & 51.00 $\pm$ 3.11 & 58.63 $\pm$ 0.37 & 53.17 $\pm$ 1.20 &- &52.59 $\pm$ 0.97 & 50.71 $\pm$ 0.76 & 54.28 $\pm$ 2.87 &-&\underline{59.52 $\pm$ 0.54}&\textbf{78.78 ± 0.06}\\
& UN Trade    & 59.65 $\pm$ 0.77 & 57.02 $\pm$ 0.69 & 61.03 $\pm$ 0.18 & 58.31 $\pm$ 3.15 & 65.24 $\pm$ 0.21 & -& 62.21 $\pm$ 0.12 & 62.17 $\pm$ 0.31 & 64.55 $\pm$ 0.62&-&\underline{65.87±0.40}&\textbf{97.39 ± 0.02}\\
\bottomrule
\end{tabular}%
}\label{cont_dynamicinap}
\end{table*}

In addition to prediction performance, we also highlight DyGSSM’s advantages in model efficiency.  Figures \ref{fig:paramcomparisiondisc} and \ref{fig:paramcomparisioncont} (see  Appendix~\ref{modelscalability}) compare model parameter sizes across different datasets on a logarithmic scale. DyGSSM consistently has the smallest parameter sizes compared to other SOTA models.   

We also performed complexity analysis of proposed $\mathcal{RW}$ for both discrete and continuous datasets (see Table~\ref{tab:runtime_cachingd},\ref{tab:runtime_cachingc} in Appendix~\ref{modelscalability}). The proposed caching mechanism yields consistent speedups between 1.8× and 2.4×, with a maximum of 3.9× improvement on large-scale graphs, corresponding to an average 50–60\% reduction in random-walk time. The gains become more substantial on larger graphs. For example on StackOverflow dataset ($\approx$2.6M nodes, $\approx$63M edges) proposed $\mathcal{RW}$ was 2× faster.  Since only structurally updated nodes trigger recomputation, runtime scales linearly with active nodes—not total graph size—enabling real-time feasibility even on million-edge networks. Note that DyGSSM retains only $\mathcal{RW}$ information related to the most recent snapshot that indicates constant memory footprint over time. In the very high-churn snapshot, worst-case scenario—the framework smoothly degrades to the standard (non-caching) $\mathcal{RW}$ complexity, ensuring no negative impact on correctness or embedding quality, only a temporary reduction in computational gain. 

To compare the extent to which DyGSSM relies on local and global structural components, we also analyzed the temporal evolution of the learned attention weights on the UCI dataset (see  Fig.~\ref{attscore} in Appendix~\ref{ablatatt}). The average attention weights remain stable over time, with slightly higher values assigned to the local component (~0.514) compared to the global component (~0.486). This indicates that the model maintains a balanced contribution from both components, with a mild preference for local neighborhood information during the fusion process. 

\subsection{Ablation study}
To evaluate the contribution of each component in DyGSSM, we performed an ablation study utilizing the DBLP and UCI datasets. 
 Specifically, we computed MRR and Recall@10 after disabling attention, local and global embeddings, and SSM. We selected 50 and 1000 negative samples per positive edge in DBLP and UCI, respectively. Table~\ref{tab:ablation} and Table~\ref{tab:ablationuci} (see Appendix~\ref{ablatatt}) show that all components contribute to DyGSSM’s overall performance. In DBLP, removing global information led to  substantial drop in performance, whereas in UCI, local information was most critical (as the attention scores also show (see Fig.~\ref{attscore} in Appendix~\ref{ablatatt}). 
 We observed performance drop when we disabled the SSM and the attention mechanism, too. 
 Both ablation studies confirm the necessity of each component.  
 
 To explicitly evaluate the contribution of the HiPPO algorithm in generating the SSM projection matrix, we conducted an experiment on  DBLP, UCI and AS733 datasets where the SSM matrix was randomly initialized using a Gaussian distribution (see Table~\ref{tab:hippo_init} in Appendix~\ref{ablatatt}). The proposed HiPPO initialization consistently improved the model performance across three datasets of varying graph sizes under two different settings (i.e., WinGNN and HawkesGNN). These results demonstrate that HiPPO initialization provides a more structured and informative state representation, enabling the model to capture temporal dependencies more effectively and achieve superior overall performance.


\begin{table*}[ht]
\centering
\scriptsize
    \caption{Ablation results for DyGSSM on DBLP dataset.}
    \begin{tabular}{@{}lccc@{}}
        \toprule
        \textbf{Model} &\textbf{MRR} & \textbf{Recall@10} \\ \midrule
        No global information &23.99$\pm$0.0050&$58.91\pm$0.0205\\
        \bottomrule
        No local information 
        & 56.40$\pm$0.0021& 92.64$\pm$0.0010\\
        \bottomrule
        No SSM 
&61.52$\pm$0.0196&97.08$\pm$0.0120
\\
        \bottomrule
        No attention & 69.55$\pm$0.0149&99.18$\pm$0.0007\\
        \bottomrule
        DyGSSM & \textbf{79.11±0.0442}&\textbf{99.85±0.0005}\\
          \bottomrule
    \end{tabular}
    \label{tab:ablation}
\end{table*}
\section{Conclusion}\label{sec:conclusion}
In this study, we propose DyGSSM, a multi-view dynamic graph representation learning approach for link prediction tasks. We trained DyGSSM in a supervised manner, leveraging both the local and global structure of each node in each snapshot to generate two distinct node embeddings. We integrate these embeddings using a lightweight attention mechanism. To mitigate RW cost, we introduced a caching mechanism that reduce the complexity and time of running RW on each time steps. To effectively incorporate past information when updating the model parameters, and to avoid the need for numerous hyperparameters, we utilized an SSM-based approach using the HiPPO algorithm to incorporate a meta-learning strategy into DyGSSM.  Experiments on 12 public datasets with two training settings show that DyGSSM outperforms SOTA models in 32 out of 36 evaluation metrics. As future work, 
 we plan to extend DyGSSM to downstream tasks such as node classification to further validate its representational power on real-world dynamic graphs. In addition, we aim to optimize the $\mathcal{RW}$  component through parallelized mini-batch sampling, graph partitioning, and subgraph-based $\mathcal{RW}$ to improve scalability on large and dense networks. Finally, we intend to explore replacing the $\mathcal{RW}$ process with learned graph embeddings. 
\bibliographystyle{unsrtnat}
\bibliography{reference}

\appendix\label{resultsappendi}

\section{Experimental Details}
\subsection{Dataset Description}\label{appdata}
We evaluated DyGSSM on discrete and continuous publicly available benchmarks.

\textit{\textbf{Bitcoin-OTC}} and \textit{\textbf{Bitcoin-Alpha}} are who-trusts-whom network,  representing trust relationships among users trading Bitcoin on Bitcoin OTC and Bitcoin Alpha platform \cite{kumar2018rev2,kumar2016edge}. These two datasets have the highest number of snapshots among all five datasets, despite having the lowest number of edges—35,592 and 24,186, respectively. 

\textit{\textbf{UCI-Message}} consists of private message communication  exchanged between students at the University of California, Irvine \cite{panzarasa2009patterns}.  It has the fewest nodes among all datasets but ranks among the top three in terms of edge density, with 59,835 edges. 

\textit{\textbf{DBLP}} represents a comprehensive list of research papers in computer science. The dataset show research collaborations between two authors, where two authors are connected if they have co-authored at least one paper \cite{hu2018developing}. 
 Note that we obtained the dynamic DBLP dataset from the WinGNN authors.
 
\textit{\textbf{Reddit-Title}} dataset consists of a hyperlink network that captures directed connections between subreddits based on hyperlinks embedded in posts linking from one subreddit to another \cite{kumar2018community}.   
 
\textit{\textbf{Reddit-Body}} captures networks of hyperlinks between subreddits, where the hyperlinks appear in the body of the posts.

\textit{\textbf{SBM}} short for Stochastic Block Model, is a widely adopted random graph model designed to simulate the evolution of community structures. 

\textit{\textbf{AS}} short for Autonomous Systems represent a communication network between routers. In this network, the nodes are routers, where each node represents a network or an AS. The edges indicate that two routers exchange traffic or routing information. 

\textit{\textbf{StackOverflow}} is a dataset containing interactions on the Stack Overflow platform. In this dataset, the nodes are users, and an edge between two users appears if one user answered another user’s question. 

\textit{\textbf{Enron}} is an email communication network from the Enron Energy Corporation. The dataset was collected over a period of three years. In this network, nodes represent email addresses, and an edge exists from address i to address j if i sent at least one email to j. 

\textit{\textbf{Can.Parl.}} is a political network that shows interactions between Canadian Members of Parliament (MPs). In this dataset, the nodes are MPs from electoral districts, and an edge is formed between two MPs when they both vote “yes” on the same bill. 

\textit{\textbf{USLegis}} dataset is a Senate co-sponsorship network that shows social interactions between legislators in the U.S. The nodes represent senators, and an edge between two nodes indicates how many times those senators co-sponsored a bill together during a given congressional session.

\textit{\textbf{UNTrade}} is a dataset of food and agriculture trade between 181 countries over the past 30 years. In this dataset, the nodes represent countries, and an edge between two nodes shows the total imports and exports of food and agricultural products exchanged between those countries. A summary of dataset statistics is presented in Tables~\ref {table:dataset}, \ref{HawkesGNNsettings}, and \ref{tab:dataset_stats}. 

\begin{table}[ht]
\caption{Dataset statistics for WinGNN settings.}
\centering 
\begin{tabular}{c c c c c}
\hline
Dataset & \#Nodes & \#Edges & \# Snapshots&Avg. Density \\ 
\hline 
Bitcoin-Alpha & 3,783 & 24,186 & 226 & 2.5890  $ \times 10^{-3}$\\ 
Bitcoin-OTC & 5,881 & 35,592 & 262 &  1.7396  $\times 10^{-3}$\\
DBLP & 28,086 & 162,451 & 27 &  9.5423  $\times 10^{-3}$\\
Reddit-Title &54,075&571,927&178& 1.9592$\times 10^{-5}$\\
UCI & 1,899 & 59,835 & 28& 1.1191 $\times 10^{-3}$\\
\hline 
\end{tabular}
\label{table:dataset} 
\end{table}

\begin{table}[ht]
\centering
\caption{Summary of dataset statistics for HawkesGNN settings.}
\begin{tabular}{lrrrrr}
\hline
Dataset & \#Nodes & \#Edges & Time Steps (Train/Val/Test) & Avg. Degree \\
\hline
UCI     & 1,899     & 59,835      & 35/5/10   & 0.36   \\
Bitcoin-Alpha   & 3,777     & 24,173      & 95/13/28  & 0.04   \\
Bitcoin-OTC     & 5,881     & 35,588      & 95/14/28  & 0.05   \\
Reddit-Title   & 54,075    & 571,927     & 122/35/17 & 0.06   \\
Reddit-Body    & 35,776    & 286,562     & 122/35/17 & 0.05   \\
AS733   & 7,716     & 1,167,892   & 70/10/20  & 2.12   \\
SBM     & 1,000     & 4,870,863   & 35/5/10   & 97.42  \\
StackOverflow      & 2,601,997 & 63,497,050  & 65/9/18   & 0.12   \\
\hline
\end{tabular}\label{HawkesGNNsettings}
\end{table}


\begin{table}[!ht]
\centering
\caption{Statistics of the datasets for continuous dynamic graph.}
\label{tab:dataset_stats}
\begin{tabular}{l l r r l l l }
\hline
Dataset & Domain & \#Nodes & \#Links & Duration & Timestamps\\
\hline
Enron & Social & 184 & 125,235 & 3 years & 22,632  \\
UCI & Social & 1,899 & 59,835  & 196 days & 58,911  \\
Can.Parl. & Politics & 734 & 74,478 & 14 years & 14  \\
USLegis. & Politics & 225 & 60,396 & 12 congresses & 12 \\
UNTrade & Economics & 255 & 507,497 & 32 years & 32 \\
\hline
\end{tabular}
\end{table} 

\subsection{Description of Baselines}\label{appbaseline}
We compare DyGSSM against state-of-the-art models on both discrete-time and continuous-time dynamic graphs.

\textit{\textbf{EvolveGCN}} \cite{pareja2020evolvegcn} introduced a recurrent mechanism to update the network parameters. In other words, it uses GCN to extract the local structure of each snapshot and injects the recurrent neural network (RNN) to capture the dynamism within the parameters of the GCN. In this study, we show the results of EvolveGCN with different temporal encoders (i.e., LSTM vs. GRU) and refer to them as EvolveGCN-O and EvolveGCN-H.


\textit{\textbf{DGNN}} \cite{manessi2020dynamic} combined GCN and LSTM to exploit both structured data and temporal information. In their study, they used stack encoder (e.g., LSTM) to capture the dynamics of nodes. 

\textit{\textbf{Dyngraph2vec}} \cite{goyal2020dyngraph2vec} used an encoder-decoder architecture to learn temporal transition in a dynamic graph. They proposed three different settings for their encoder-decoder architectures, composed of dense and recurrent based models. 

\textit{\textbf{ROLAND}} \cite{you2022roland} is a meta-learning based approach that update the model parameters of the adjacent snapshots. They introduced a live update based mechanisem on the traditional GNN layer, that makes their model adoptable to convert static graph to dynamic graph learning. 

\textit{\textbf{WinGNN}} \cite{zhu2023wingnn} is another meta-learning method that introduces an encoder-free architecture to extract the dynamics.  

\textit{\textbf{TransformerG2G}} \cite{varghese2024transformerg2g} is a transformer based model that aim to obtain lower-dimensional multivariate Gaussian representations of nodes, that effectively capture long-term temporal dynamics. They  trained the transformer encoder from the second timesteps when  weights transferred from the pre-trained model for the first
graph snapshot embedding.  

\textit{\textbf{DTFormer}} \cite{chen2024dtformer} is another transformer based model. They used attention mechanism to capture topological information in each time steps and temporal dynamics of graphs along the timestamps. 

\textit{\textbf{DG-Mamba}} \cite{yuan2024dg} is a SSM based method that design to extract long dependency on dynamic graph. The authors introduced kernelized dynamic message-passing operator. To capture global intrinsic dynamics, we establish the dynamic graph as a self-contained system with SSM.  

\textit{\textbf{DySAT}} \cite{sankar2020dysat} learns node representations by jointly applying self-attention across structural neighborhoods and temporal dynamics to capture both relational structure and temporal evolution.

\textit{\textbf{VGRNN}} \cite{hajiramezanali2019variational} is a hierarchical variational model that introduces latent random variables to  jointly captures both topology and node attribute changes in dynamic graphs. 

\textit{\textbf{HTGN}} \cite{yang2021discrete} is a model that captures how networks evolve over time by embedding them in hyperbolic space. It uses hyperbolic GNNs, recurrent units, attention, and a stability module to learn evolving patterns effectively and reliably.

\textit{\textbf{M2DNE}} \cite{lu2019temporal} is a temporal network embedding method that models both micro- and macro-dynamics of evolving networks. It uses a temporal attention to capture fine-grained edge events and a dynamics equation to enforce higher-level structural evolution in node embeddings.

\textit{\textbf{GHP}} \cite{shang2019geometric} integrates Hawkes processes with a graph convolutional recurrent neural network. It is also computationally efficient, using a constant number of parameters regardless of graph size. 

\textit{\textbf{HawkesGNN}} fused multiple snapshots into a single temporal graph by combining Hawkes process with GNN. They used a Hawkes excitation matrix to model the temporal edges.

\textit{\textbf{JODIE}} \cite{kumar2019predicting} is a coupled recurrent neural network that learns user and item embedding trajectories, predicts future embeddings via a novel projection operator, and accelerates training using the scalable t-Batch algorithm. 

\textit{\textbf{DyRep}} \cite{trivedi2019dyrep} is an inductive deep learning framework that generates low-dimensional node embeddings evolving over time. It models the communication and association dynamics between nodes using a time-scale-dependent multivariate point process. 

\textit{\textbf{TGAT}} \cite{xu2020inductive} is a temporal graph attention layer that aggregates temporal and topological neighborhood features using self-attention and a functional time encoding based on Bochner’s theorem

\textit{\textbf{TGN}} \cite{rossi2020temporal} combines memory modules with graph-based operators to achieve superior performance, using a message function, message aggregator, and memory updater.

\textit{\textbf{CAWN}} \cite{wang2021inductive} captures network dynamics through temporal random walks. CAWs anonymize node identities using hitting counts to maintain inductiveness and motif correlations. These are then encoded by the CAW-N neural network, paired with a constant-time, and constant-memory sampling strategy.

\textit{\textbf{EdgeBank}} \cite{poursafaei2022towards} is a memory-based baseline for dynamic link prediction that stores past interactions and predicts edges as positive if observed. It has four variants—unlimited memory, fixed time-window (two versions), and threshold-based, which allow flexible memory management.

\textit{\textbf{TCL}} \cite{wang2021tcl} TCL is a graph neural network for continuous-time dynamic graphs. It introduces a graph-topology-aware Transformer, a two-stream encoder with co-attentional modeling of interaction dependencies, and a contrastive learning objective that maximizes mutual information between future interaction nodes.

\textit{\textbf{GraphMixer}} \cite{cong2023we} is a simple yet effective architecture composed of three components: an MLP-based link encoder, a neighbor mean-pooling node encoder, and an MLP-based link classifier.

\textit{\textbf{DyGFormer}}  \cite{yu2023towards} is a Transformer-based model that only relies on nodes’ historical first-hop interactions. It encodes neighbor co-occurrences to capture source–destination correlations and uses a patching technique to handle longer histories.

\textit{\textbf{FreeDyG}} \cite{tian2024freedyg} is a continuous-time dynamic graph model for link prediction that enhances learning by encoding node interaction frequency. Unlike prior time-domain methods, it leverages the frequency domain to capture periodic and shifting interaction patterns.

\begin{figure*}[ht]
    \centering
    \includegraphics[width=14cm]{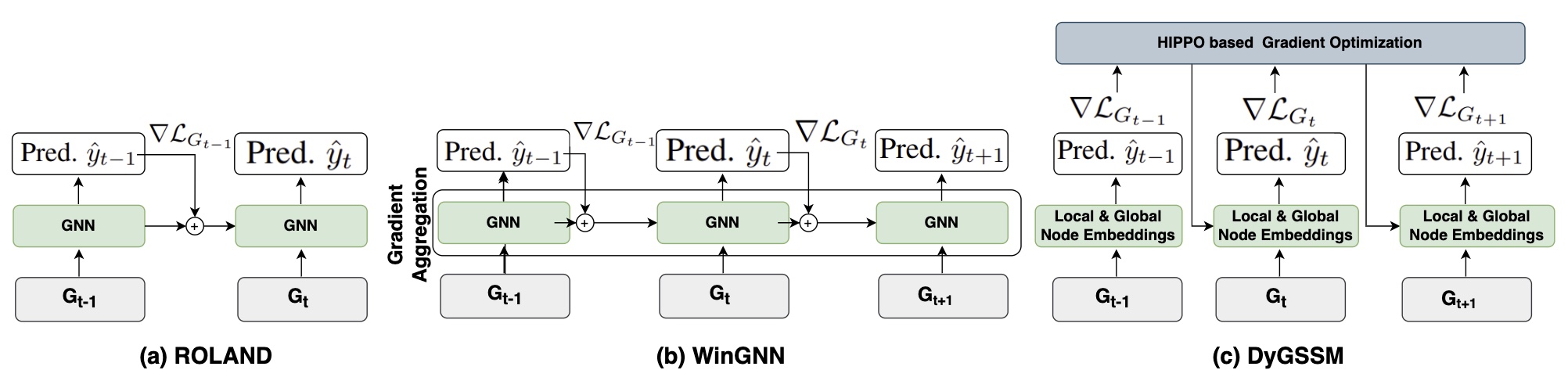}
    \caption{The figure compares how ROLAND, WinGNN, and DyGSSM update their model parameters.
(a) ROLAND updates model parameters between consecutive time steps using fixed meta-learning weights.
(b) WinGNN updates parameters between time steps with a fixed learning rate. Instead of relying on explicit time encoding, it uses a window-based gradient aggregation mechanism.
(c) DyGSSM utilizes the HiPPO-based algorithm to update model parameters without a need to specify a window size.}
    \label{fig:enter-arc}
\end{figure*}

\section{Evaluation Metrics and Implementation Details}
\label{app-eval-metrics}
We evaluate the effectiveness of our model using four widely adopted metrics: accuracy, macro-AUC, Mean Reciprocal Rank (MRR), and Recall@10. Among these, MRR and Recall@10 are our primary evaluation metrics, as accuracy and AUC can be overly sensitive to imbalanced class distributions. To ensure a fair comparison with SOTA methods, we follow ROLAND \cite{you2022roland} framework for future link prediction task. For each node $u$ with a positive edge ($u$,$v$) at time $t+1$, we randomly sample 1,000 negative edges originating from $u$. The rank of the prediction score for the positive edge ($u$,$v$) is then determined relative to the scores of the sampled negative edges. The MRR is computed as the average of the reciprocal ranks across all nodes $u$. Using the same ranking, Recall@10 is calculated as the proportion of positive edges ranked within the top 10. It is worth noting that, due to memory constraints, we limit the sampling to 50 negative edges for DTFormer and DG-Mamba on the DBLP dataset, as indicated in Table \ref{tab:reddit_title_metrics} with an asterisk. 
To have a fair comparison with the HawkesGNN model, we used Average Precision (as they used in their paper) to compare DyGSSM with the SOTA model. We used their source code
and integrated DyGSSM into their code. 
All the results in Table \ref{tab:reddit_title_metrics} up to WinGNN columns are coming from the WinGNN paper \cite{zhu2023wingnn}. We followed WinGNN in train test data division, 70\%  of snapshots for training, and remaining 30\%  for testing. Training is set for 100 epochs, with patience of 10 epochs for early stopping. We used Adam as our optimizer and repeated the experiment with 10 random seeds to ensure robust
error estimation.  All the results in Table \ref{tab:overall_performance_HawkesGNN} come from the HawkesGNN paper. All experiments are performed on a single
GPU equipped with Nvidia A100 with 80GB of memory. 

\section{Additional Results}\label{resultsappendix}

\begin{table*}[!ht]
    \caption{Average Precision (AP) score comparison on five datasets. The best and second best results are shown in bold and \underline{underlined}, respectively. We repeated the experiment with 10 random seeds and reported the average metrics with standard deviation.}
     \resizebox{\textwidth}{!}{%
    \begin{tabular}{@{}ccccccccc@{}}
        \toprule
         \textbf{Dataset} &\textbf{WinGNN}&\textbf{TransformerG2G}&\textbf{DTFormer} &\textbf{DG-Mamba}& \textbf{DyGSSM}\\ \midrule
         DBLP& \underline{92.96±0.0019}&59.41$\pm$0.0077& 82.03±0.0119 &53.70±0.0385& \textbf{98.69$\pm$0.0002}\\
        UCI&\underline{96.49±0.0119}&74.36$\pm$0.0660& 86.19±0.0208 &72.54±0.0063& \textbf{98.89$\pm$0.0061}\\
        Bitcoin-OTC&92.25$\pm$0.0067&63.25$\pm$0.0518&\textbf{97.83$\pm$0.0039}&OOM&\underline{95.52$\pm$0.0284}\\
        Bitcoin-Alpha&93.85$\pm$0.0139&OOM&\underline{95.97±0.0162}&OOM&\textbf{97.29$\pm$0.0229}\\
        Reddit-Title&\underline{99.99$\pm$0.0001}&OOM&95.80$\pm$0.0000&OOM&\textbf{99.99$\pm$0.0000}\\
        \bottomrule
    \end{tabular}}
    \label{tab:ap}
\end{table*}

\begin{table*}[!ht]
    \caption{Performance comparison of MRR and Recall@10 on four datasets for the models with  asterisk  in Table \ref{tab:reddit_title_metrics} using 50 negative samples instead of 1000. The best and second best results are shown in bold and \underline{underlined}, respectively. We repeated the experiment with 10 random seeds and reported the average metrics with standard deviation. TransformerG2G for DBLP results are not shown as they are available in Table \ref{tab:reddit_title_metrics}. OOM: out-of-memory}
    \begin{tabular}{@{}lcccccc@{}}
        \toprule
        \textbf{Dataset} & \textbf{Metric} &\textbf{DTFormer}&\textbf{DG-Mamba}&\textbf{TransformerG2G}&\textbf{DyGSSM}  \\ \midrule
        \multirow{2}{*}{DBLP} 
        & MRR& \underline{61.07±0.0102}& 16.00±0.0042&-&\textbf{79.11±0.0442}  \\
        & Recall@10&\underline{68.14±0.01658}&38.71±0.0230&-&\textbf{99.85±0.0005}
        \\ \bottomrule
        \multirow{2}{*}{Bitcoin-OTC}
        & MRR&\textbf{77.49$\pm$0.0266}&OOM&48.40$\pm$0.1327&\underline{75.35$\pm$0.0802}\\
        & Recall@10&\textbf{85.95$\pm$0.0225}&OOM&60.72$\pm$0.1618&\underline{78.21$\pm$0.0987}
        \\ \bottomrule
        \multirow{2}{*}{Bitcoin-Alpha}
        & MRR&\underline{55.22±0.0307}&OOM&OOM&\textbf{77.88$\pm$0.0449} \\
        & Recall@10&\underline{71.21±0.0532}&OOM&OOM&\textbf{79.33$\pm$0.0450}\\ \bottomrule
        \multirow{2}{*}{Reddit-Title}
        & MRR&\underline {80.12$\pm$0.0054}&OOM&OOM&\textbf{96.43$\pm$0.0067} \\
        & Recall@10&\underline{85.94$\pm$0.0006}&OOM&OOM& \textbf{99.99$\pm$0.0000}
        \\ \bottomrule
    \end{tabular}
    \label{tab:50mrr}
\end{table*}

\begin{table*}[!ht]
\centering
\caption{AUC-ROC for transductive and inductive dynamic link prediction with random negative sampling strategies. The best and second best results are shown in \textbf{bold} and \underline{underlined}, respectively. The results are taken from the DyGFormer and FreeDyG papers. Since the FreeDyG authors did not evaluate their model on Can. Parl, US Legist, and UN Trade, we used the results reported for DyGFormer on these datasets and marked FreeDyG with “–”. Inductive results for EdgeBank were not reported by either DyGFormer or FreeDyG and are marked as “–”.}
\resizebox{\textwidth}{!}{%
\begin{tabular}{l|l|ccccccccccccc}
\toprule
\textbf{Settings} & \textbf{Datasets} & \textbf{JODIE} & \textbf{DyRep} & \textbf{TGAT} & \textbf{TGN} & \textbf{CAWN} & \textbf{EdgeBank} & \textbf{TCL} & \textbf{GraphMixer} & \textbf{DyGFormer} & \textbf{FreeDyG} & \textbf{DyG-Mamba} & \textbf{DyGSSM} \\
\midrule
\multirow{5}{*}{\rotatebox[origin=c]{90}{Transductive}} 
& Enron     & 87.96$\pm$0.52 & 84.89$\pm$3.00 & 68.89$\pm$1.10 & 88.32$\pm$0.99 & 90.45$\pm$0.14 & 87.05$\pm$0.00 & 75.74$\pm$0.72 & 84.38$\pm$0.21 & \underline{93.33$\pm$0.13} & \textbf{94.01$\pm$0.11} & 93.05$\pm$0.17 & 92.60$\pm$0.01 \\
& UCI       & 90.44$\pm$0.49 & 68.77$\pm$2.34 & 78.53$\pm$0.74 & 92.03$\pm$1.13 & 93.87$\pm$0.08 & 77.30$\pm$0.00 & 87.82$\pm$1.36 & 91.81$\pm$0.67 & 94.49$\pm$0.26 & 95.00$\pm$0.21 & \underline{95.32$\pm$0.18} & \textbf{96.95$\pm$0.00} \\
& Can. Parl. & 78.21$\pm$0.23 & 73.35$\pm$3.67 & 75.69$\pm$0.78 & 76.99$\pm$1.80 & 75.70$\pm$3.27 & 64.14$\pm$0.00 & 72.46$\pm$3.23 & 83.17$\pm$0.53 & 97.76$\pm$0.41 & – & \underline{98.67$\pm$0.29} & \textbf{99.99$\pm$0.00} \\
& US Legis.  & 82.85$\pm$1.07 & 82.28$\pm$0.32 & 75.84$\pm$1.99 & 83.34$\pm$0.43 & 77.16$\pm$0.39 & 62.57$\pm$0.00 & 76.27$\pm$0.63 & 76.96$\pm$0.79 & 77.90$\pm$0.58 & – & \underline{78.19$\pm$0.64} & \textbf{92.86$\pm$0.03} \\
& UN Trade   & 69.62$\pm$0.44 & 67.44$\pm$0.83 & 64.01$\pm$0.12 & 69.10$\pm$1.67 & 68.54$\pm$0.18 & 66.75$\pm$0.00 & 64.72$\pm$0.05 & 65.52$\pm$0.51 & 70.20$\pm$1.44 & – & \underline{72.19$\pm$0.09} & \textbf{99.99$\pm$0.00} \\
\midrule
\multirow{5}{*}{\rotatebox[origin=c]{90}{Inductive}} 
& Enron      & 81.96$\pm$1.34 & 76.34$\pm$4.20 & 64.63$\pm$1.74 & 78.83$\pm$1.11 & 87.02$\pm$0.50 & – & 72.33$\pm$0.99 & 76.51$\pm$0.71 & 90.69$\pm$0.26 & 89.51$\pm$0.20 & \underline{90.84$\pm$0.18} & \textbf{97.40$\pm$0.01} \\
& UCI        & 78.80$\pm$0.94 & 58.08$\pm$1.81 & 77.64$\pm$0.38 & 86.68$\pm$2.29 & 90.40$\pm$0.11 & – & 84.49$\pm$1.82 & 89.30$\pm$0.57 & 92.63$\pm$0.13 & \underline{93.01$\pm$0.08} & 91.99$\pm$0.03 & \textbf{96.77$\pm$0.00} \\
& Can. Parl. & 53.81$\pm$1.14 & 55.27$\pm$0.49 & 56.51$\pm$0.75 & 55.86$\pm$0.75 & 58.83$\pm$1.13 & – & 55.83$\pm$1.07 & 58.32$\pm$1.08 & 89.33$\pm$0.48 & – & \underline{90.77$\pm$0.86} & \textbf{99.99$\pm$0.00} \\
& US Legis.  & 58.12$\pm$2.35 & 61.07$\pm$0.56 & 48.27$\pm$3.50 & 62.38$\pm$0.48 & 51.49$\pm$1.13 & – & 50.43$\pm$1.48 & 47.20$\pm$0.89 & 53.21$\pm$3.04 & – & \underline{56.56$\pm$1.08} & \textbf{74.35$\pm$0.07} \\
& UN Trade   & 62.28$\pm$0.50 & 58.82$\pm$0.98 & 62.72$\pm$0.12 & 59.99$\pm$3.50 & 67.05$\pm$0.21 & – & 63.76$\pm$0.07 & 63.48$\pm$0.37 & 67.25$\pm$1.05 & – & \underline{69.22$\pm$0.52} & \textbf{96.59$\pm$0.02} \\
\bottomrule
\end{tabular}%
}
\label{tab:auc_merged}
\end{table*}

\newpage

\section{Model Scalability}\label{modelscalability}
\begin{figure}[!ht]
    \centering
    \includegraphics[width=0.7\linewidth]{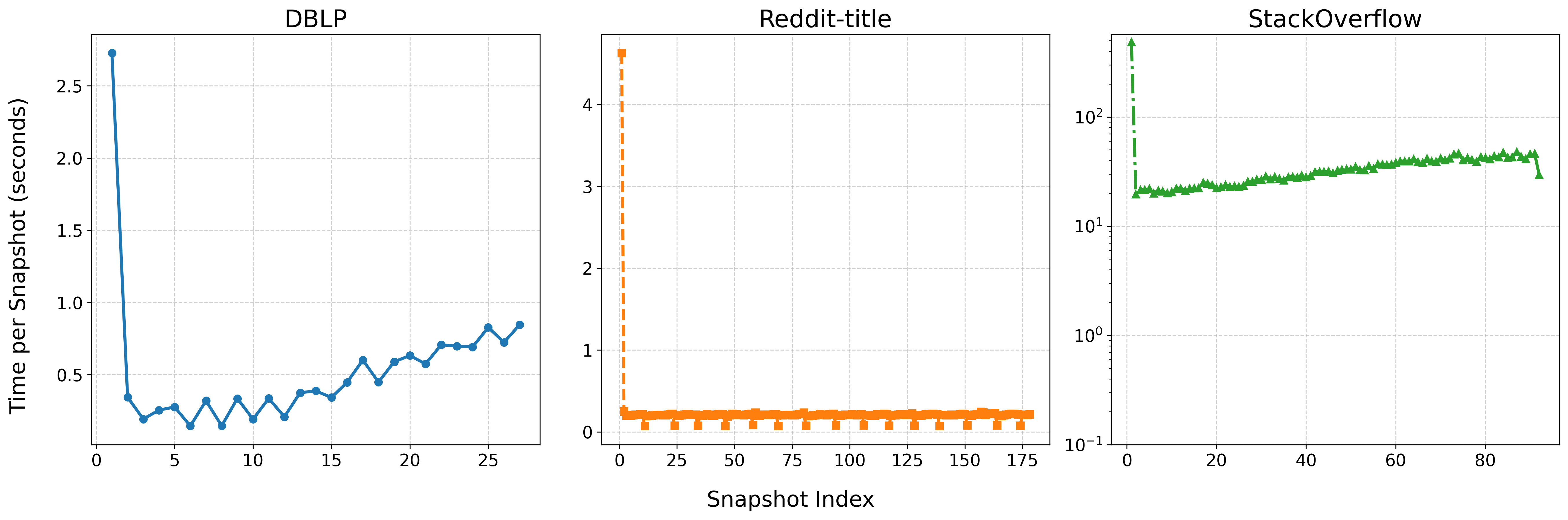}
    \caption{Global neighborhood computation time using RW on DBLP, Reddit-Title, and StackOverflow datasets after applying the caching mechanism. The computation cost per snapshot is initially high, but it significantly decreases when the caching mechanism is used. 
    }
    \label{fig:snapshot_times_scaled}
\end{figure}

\begin{figure}[!ht]
    \centering
    \includegraphics[width=0.7\linewidth]{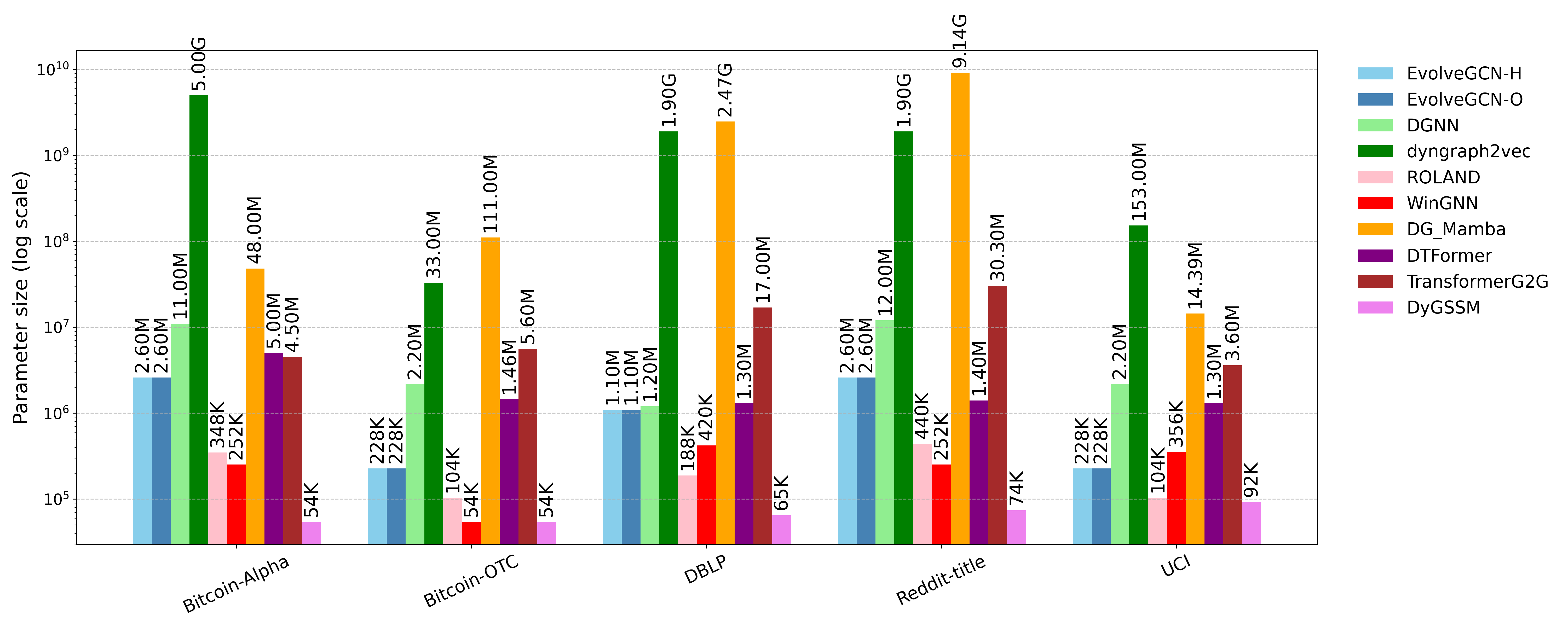}
    \caption{Model parameter size comparison discrete-time dynamic graphs. Each bar represents a model and its number of learnable parameters in millions (M) or thousands (K). DyGSSM consistently has one of the smallest parameter sizes, typically ranging from 50K to 92K. Despite integrating GCN, Conv1D, and light attention, our model remains lightweight and highly scalable.}
    \label{fig:paramcomparisiondisc}
\end{figure}
\begin{figure}[!ht]
    \centering
    \includegraphics[width=0.6\linewidth]{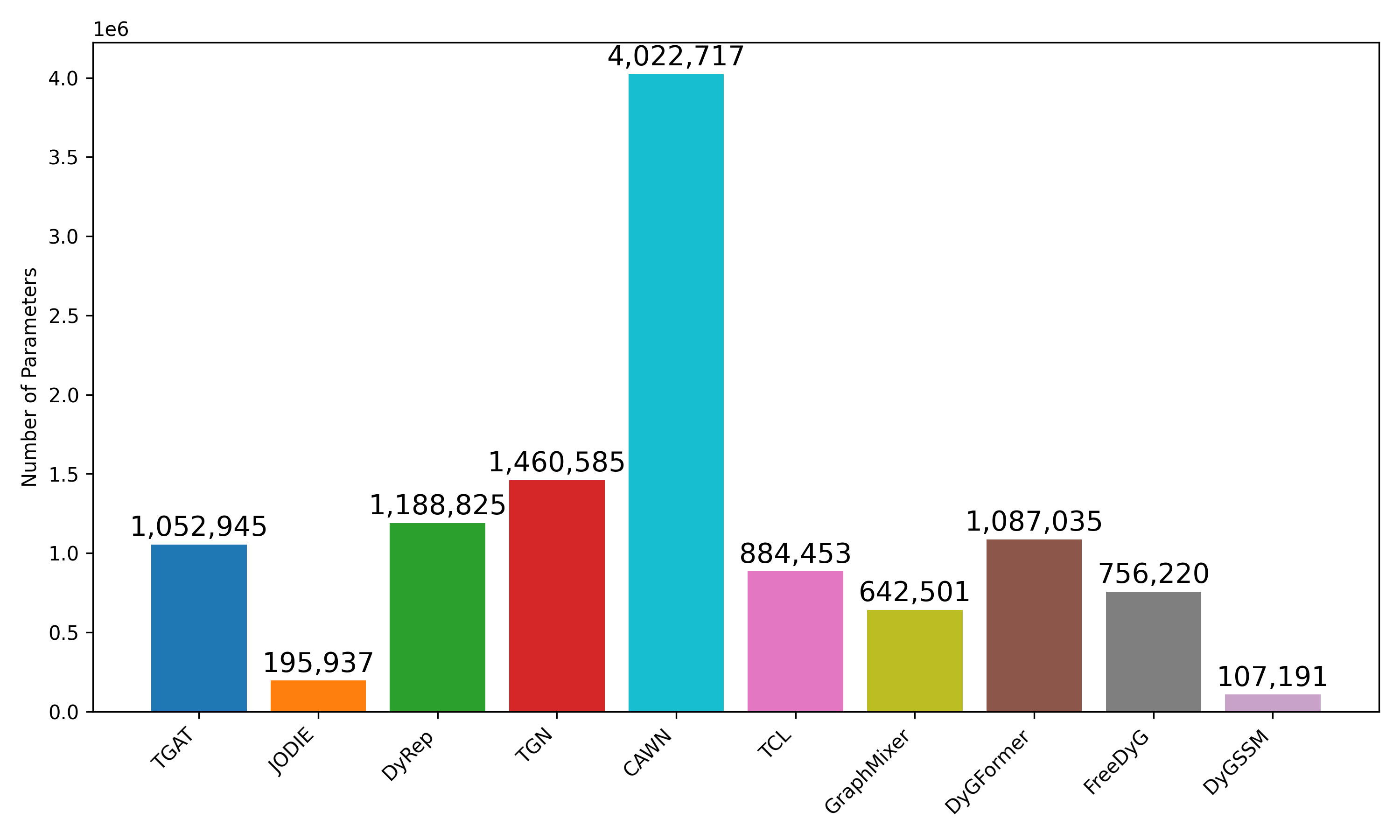}
    \caption{Model parameter size comparison (continuous-time dynamic graph). Each bar represents a model and its number of learnable parameters in millions (M) or thousands (K). DyGSSM has the smallest parameter size.}
    \label{fig:paramcomparisioncont}
\end{figure}


\begin{table}[!ht]
\centering
\caption{Runtime comparison of the random-walk (RW) component with and without caching across seven discrete dynamic graph datasets.}
{\scriptsize
\begin{tabular}{@{}lccrrrr@{}}
\toprule
\textbf{Dataset} & \textbf{\# Nodes} & \textbf{\# Edges} & \textbf{Normal RW (s)} & \textbf{Caching RW (s)} & \textbf{Speedup (× Faster)} & \textbf{\% Time Reduction} \\
\midrule
Bitcoin-OTC     & 5,881     & 35,588      & 5.31   & 2.67   & 1.99× & 49.8\% \\
Bitcoin-Alpha   & 3,777     & 24,173      & 3.63   & 1.50   & 2.42× & 58.8\% \\
UCI             & 1,899     & 59,835      & 1.34   & 0.70   & 1.91× & 47.8\% \\
Reddit-Title    & 54,075    & 571,927     & 61.86  & 34.26  & 1.81× & 44.6\% \\
Reddit-Body     & 35,776    & 286,562     & 41.61  & 21.90  & 1.90× & 47.4\% \\
AS733           & 7,716     & 1,167,892   & 6.13   & 6.01   & 1.02× & $\sim$2\% \\
Stack Overflow  & 2,601,997 & 63,497,050  & 1795.54 & 897.01 & 2.00× & 50.0\% \\
\bottomrule
\end{tabular}
}
\label{tab:runtime_cachingd}
\end{table}

\begin{table}[!ht]
\centering
\caption{Runtime comparison of the random walk with and without caching across four continuous dynamic graph datasets.}
{\scriptsize
\begin{tabular}{@{}lccccc@{}}
\toprule
\textbf{Dataset} & \textbf{\# Nodes} & \textbf{\# Edges} & \textbf{Without Caching} & \textbf{With Caching} & \textbf{Speedup (×) / \% Time Reduction} \\
\midrule
Reddit   & 10,984 & 672,447  & 5 min 59 s ($\approx$ 359 s) & 1 min 32 s ($\approx$ 92 s)  & 3.9× / 74\% \\
UN Vote  & 201    & 1,035,742 & 14 min 28 s ($\approx$ 868 s) & 7 min 17 s ($\approx$ 437 s) & 2.0× / 50\% \\
UN Trade & 255    & 507,497  & 3 min 12 s ($\approx$ 192 s) & 0 min 52 s ($\approx$ 52 s)  & 3.7× / 73\% \\
\bottomrule
\end{tabular}
}
\label{tab:runtime_cachingc}
\end{table}

\newpage

\section{Ablation and Attention Results}\label{ablatatt}

\begin{table*}[!ht]
\centering
\scriptsize
    \caption{Ablation results for DyGSSM on UCI dataset.}
    \begin{tabular}{@{}lccc@{}}
        \toprule
        \textbf{Model} &\textbf{MRR} & \textbf{Recall@10} \\ \midrule
        No global information &\underline{30.85$\pm$0.0011}&\underline{55.17$\pm$0.0031}\\
        \bottomrule
        No local information 
        &20.11$\pm$0.0000& 45.40$\pm$0.0000\\
        \bottomrule
        No SSM & 27.32$\pm$0.0036& 55.05$\pm$0.0083\\
        \bottomrule
        No attention &25.57$\pm$0.0131&54.76$\pm$0.0012\\
        \bottomrule
        DyGSSM &\textbf{42.92$\pm$0.0072}
&\textbf{74.08$\pm$0.0018}
\\
          \bottomrule
    \end{tabular}
    \label{tab:ablationuci}
\end{table*}

\begin{table}[!ht]
\centering
\caption{Comparison of Random and HiPPO-based initialization on the DBLP dataset.}
{\scriptsize
\begin{tabular}{@{}llcccc@{}}
\toprule
\textbf{Dataset} & \textbf{Initialization Setting} & \textbf{MRR (↑)}  & \textbf{\# Nodes} & \textbf{\# Edges} \\
\midrule
DBLP & Gaussian distribution & 19.18 $\pm$ 0.0026 &  28,086 & 162,451 \\
 & \textbf{HiPPO} & \textbf{27.90 $\pm$ 0.0449} &   &  \\
\midrule
UCI   & Gaussian distribution & 33.59 $\pm$ 0.0081 & 1,899 & 59,835 \\
   & \textbf{HiPPO} & \textbf{36.08 $\pm$ 0.0300} &  &  \\
\midrule
AS733 & Gaussian distribution & 38.14 $\pm$ 0.0222 & 7,716 & 1,167,892 \\
 & \textbf{HiPPO} & \textbf{52.64 $\pm$ 0.0000} &  &  \\
\bottomrule
\end{tabular}
}
\label{tab:hippo_init}
\end{table}

\begin{figure}[!ht]
    \centering
    \includegraphics[width=0.6\linewidth]{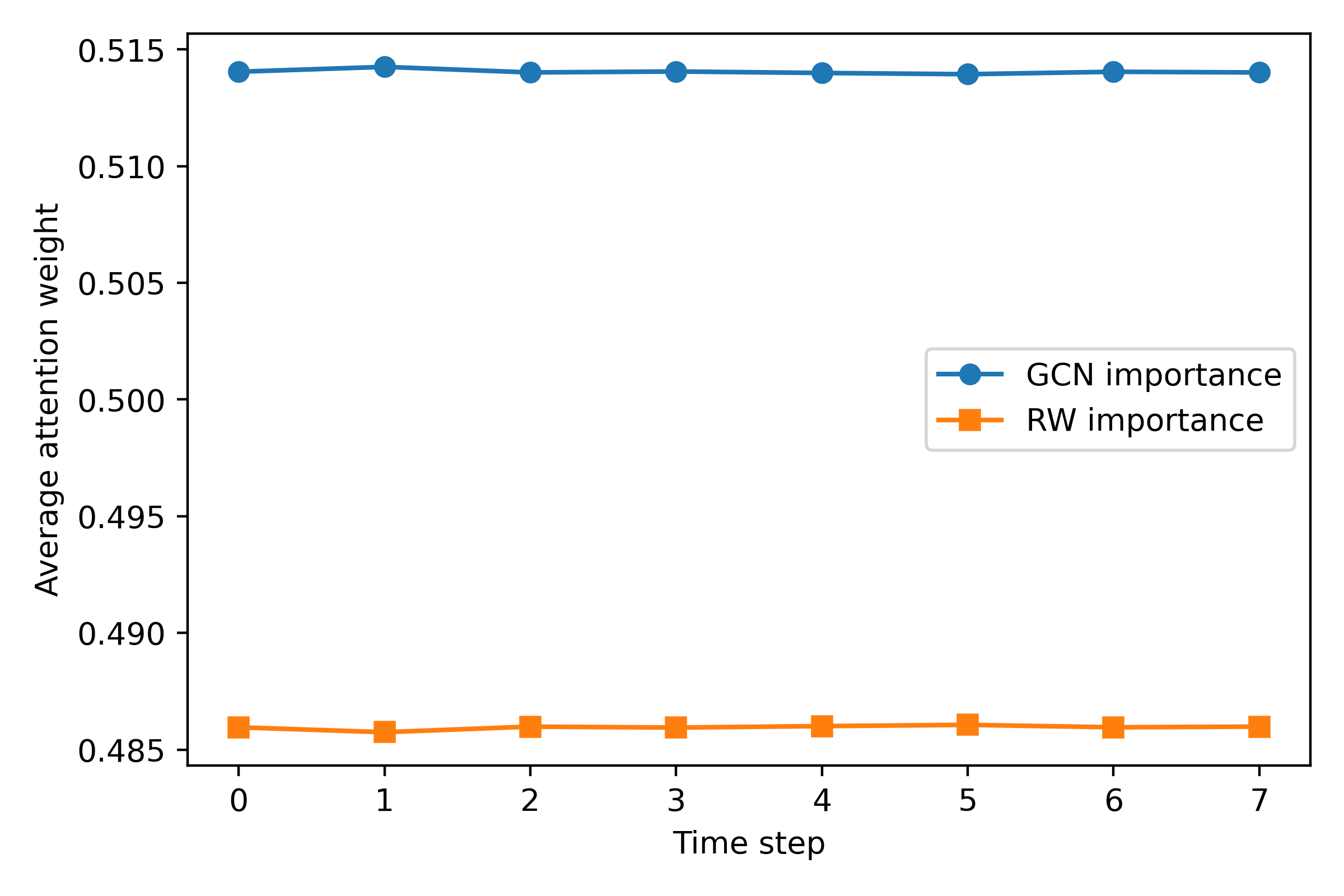}
    \caption{DyGSSM attention to local and global structure on UCI dataset.}
    \label{attscore}
\end{figure}
\end{document}